\RequirePackage{fix-cm}
\documentclass[twocolumn]{svjour3}          % twocolumn

\smartqed  % flush right qed marks, e.g. at end of proof

\usepackage{times}
\usepackage{graphicx}
\graphicspath{{./ijcv1.1/}{./figs/}}
\DeclareGraphicsExtensions{.pdf,.jpeg,.png}
\usepackage{amsmath,amssymb,bm,bbm} 
\usepackage{amsmath,amsfonts}
\usepackage{algorithmic}
\usepackage{algorithm}
\usepackage{tikz}
\usepackage{comment}
\usepackage{color}
\usepackage{overpic}
\usepackage{graphicx}
\usepackage{pifont}
\usepackage{pict2e} 
\usepackage{booktabs}
\usepackage{multirow}
\usepackage{multicol}
\usepackage{xcolor,colortbl}
\usepackage{array}
\usepackage{rotating}

\usepackage{subcaption}
\newcommand{\myPara}[1]{\par\vspace{.0in}\noindent\textbf{#1}.\ }

\usepackage[authoryear]{natbib}

\usepackage[colorlinks,linkcolor=blue,citecolor=black]{hyperref}

\makeatletter
\let\cl@chapter\relax
\makeatother

\usepackage[capitalize,noabbrev]{cleveref}
\crefname{section}{Sec.}{Secs.}
\crefname{figure}{Fig.}{Figs.}
\crefname{table}{Tab.}{Tabs.}
\crefname{equation}{Eq.}{Eqs.}
\crefname{algorithm}{Alg.}{Algs.}

\definecolor{color1}{RGB}{237,125,49}
\definecolor{color3}{RGB}{0,112,192}
\definecolor{color2}{RGB}{112,173,71}
\usepackage{etoolbox}
\usepackage{listofitems}
\usepackage{wheelchart}
\usetikzlibrary{decorations.text}
\usetikzlibrary{decorations.markings}
\usepackage{makecell}

\usepackage{forest}
\usetikzlibrary{trees,positioning,shapes,shadows,arrows.meta}

\makeatletter
\renewcommand\subsubsection{\@startsection{subsubsection}{4}{1\parindent}{0ex plus 0.1ex minus 0.1ex}{0ex}{\normalfont\normalsize\bfseries\itshape}}\renewcommand\paragraph{\@startsection{paragraph}{4}{2\parindent}{0ex plus 0.1ex minus 0.1ex}{0ex}{\normalfont\normalsize\bfseries\itshape}}\makeatother

\usetikzlibrary{trees}
\definecolor{hidden-draw}{rgb}{0.5, 0.5, 0.5}
\definecolor{harvestgold}{rgb}{0.85, 0.57, 0.0}
\definecolor{cyan}{rgb}{0.0, 1.0, 1.0}
\definecolor{lightcoral}{rgb}{0.94, 0.5, 0.5}
\definecolor{SlateBlue}{rgb}{0.416, 0.353, 0.804}
\definecolor{LightGreen}{rgb}{0.564, 0.933, 0.564}
\definecolor{Turquoise}{rgb}{0.251, 0.878, 0.816}
\definecolor{BlueGreen}{rgb}{0.643, 0.859, 0.871}
\definecolor{BlueViolet}{rgb}{0.54, 0.17, 0.89}
\definecolor{RAG}{rgb}{0.639, 0.835, 0}
\definecolor{0}{rgb}{0.92, 0.3, 0.26}
\definecolor{1}{rgb}{0.961, 0.851, 0.471}
\definecolor{1_1}{rgb}{1.00, 0.933, 0.698}
\definecolor{2}{rgb}{0.522, 0.784, 0.953}
\definecolor{2_1}{rgb}{0.722, 0.592, 0.871}
\definecolor{2_2_2}{rgb}{0.561, 0.659, 0.843}
\definecolor{3}{rgb}{0.56, 0.93, 0.56}
\definecolor{3_1_1}{rgb}{0.808, 0.996, 0.808}
\definecolor{4}{rgb}{0.988, 0.541, 0.565}
\definecolor{4_1}{rgb}{0.996, 0.796, 0.808}
\definecolor{4_1_1}{rgb}{0.957, 0.847, 0.843}
\definecolor{bounding}{rgb}{0.855, 0.392, 0.357}
\definecolor{6}{rgb}{0.690, 0.612, 0.522}

\definecolor{golden}{RGB}{255, 215, 0}
\definecolor{golden}{RGB}{0, 0, 0}

\definecolor{backbone}{RGB}{255,192,0}
\definecolor{neck}{RGB}{112,173,71}
\definecolor{head}{RGB}{237,125,49}
\definecolor{score}{RGB}{68,72,212}
\definecolor{grade}{RGB}{92,201,207}
\definecolor{rank}{RGB}{0,176,207}

\usepackage{fontawesome}
\usepackage{calc}

\usepackage{orcidlink}
\usepackage{ragged2e}

\newlength{\myMheight}
\usepackage{MnSymbol,bbding,pifont}

\usepackage[misc]{ifsym}
\usepackage{nicematrix}

\usepackage{pdfpages}

\usepackage{siunitx}
\usepackage[pagewise]{lineno}
\usepackage{xstring}   

\colorlet{highlightcolor}{red!70!black}
\newif\ifMShighlight
\MShighlighttrue      

\newif\ifMSinputmode
\MSinputmodefalse

\newcommand{\MSfigcaption}[3]{\ifMSinputmode
    \caption{\textcolor{highlightcolor}{#2}}\else
    \caption{#2}\fi
  \label{#1}#3}

\newcommand{\MStabcaption}[2]{\ifMSinputmode
    \caption{\textcolor{highlightcolor}{#2}}\else
    \caption{#2}\fi
  \label{#1}}

\newcommand{\MSeqlabel}[1]{\label{#1}}
\newcommand{\MSlinelabel}[1]{\unskip\noindent  %\linelabel{#1}
\ignorespaces    }

\makeatletter
\let\MS@originput\input
\renewcommand{\input}[1]{\IfBeginWith{#1}{r2/}{
    \ifMShighlight
      \begingroup
        \MSinputmodetrue        \label{start:#1}\color{highlightcolor}\arrayrulecolor{highlightcolor}\MS@originput{#1}\arrayrulecolor{black}\label{end:#1}\endgroup
      \ignorespaces
    \else 
      \MS@originput{#1}\ignorespaces
    \fi
  }{\MS@originput{#1}\ignorespaces
  }}
\makeatother

\usepackage{times}

\DeclareCaptionLabelSeparator{quadsep}{\quad}

\usepackage{etoolbox}

\makeatletter
\newcounter{mycitecount}
\newcounter{mycitetotal}

\newcommand{\myciteyearlink}[1]{%
  \hyperlink{cite.#1}{\textcolor{blue}{\NoHyper\citeyear{#1}\endNoHyper}}%
}

\newcommand{\mycitep}[1]{%
  \setcounter{mycitecount}{0}%
  \setcounter{mycitetotal}{0}%
  \renewcommand*{\do}[1]{\stepcounter{mycitetotal}}%
  \docsvlist{#1}%
  (%
  \renewcommand*{\do}[1]{%
    \stepcounter{mycitecount}%
    \citeauthor{##1},~\myciteyearlink{##1}%
    \ifnum\value{mycitecount}<\value{mycitetotal};\ \fi
  }%
  \docsvlist{#1}%
  )%
}

\DeclareRobustCommand{\cite}[1]{\mycitep{#1}}
\makeatother
\makeatletter

\renewcommand{\citet}[1]{%
  \@for\@citekey:=#1\do{%
    \citeauthor{\@citekey}~(\myciteyearlink{\@citekey})%
  }%
}

\makeatother

\begin{document}
\sloppy

\title{
Two-Stage Multi-Modal Fusion with Adaptive Alignment for Action Quality Assessment
}

\titlerunning{Two-Stage Multi-Modal Fusion with Adaptive Alignment}        % if too long for running head

% \author{
%     Author 1 \and
%     Author 2
% }

% \institute{
%         Author 1 \at
%         University 1 \\
%         \email{xx@xx.edu.cn}      
%         \and
%         Author 2 \at
%         University 2 \\
%         \email{xx@xx.edu.cn}   
% }

\author{
Kanglei Zhou~\orcidlink{0000-0002-4660-581X} \and
Ruizhi Cai~\orcidlink{0009-0009-3328-0559} \and
Xinning Wang~\orcidlink{0000-0003-0254-6683} \and
Yijian Zheng~\orcidlink{0009-0000-7757-3083} \and
Liyuan Wang~\orcidlink{0009-0002-7797-325X} \and
Jianguo Li~\orcidlink{0000-0001-9431-2950} \and
Xiaohui Liang~\orcidlink{0000-0001-6351-2538}
% \thanks{Corresponding authors: Xinning Wang (xnwangcip@hotmail.com), Jianguo Li (jianguo\_li6@hotmail.com), and Xiaohui Liang (liang\_xiaohui@buaa.edu.cn).}
}

% \authorrunning{Zhou et al.} % if too long for running head

\institute{
Kanglei Zhou, Liyuan Wang \at Dept. of Psychological and Cognitive Sciences, Tsinghua University, Beijing, China \\
\email{\{zhoukanglei, liyuanwang\}@tsinghua.edu.cn}
\and
Ruizhi Cai, Yijian Zheng, Xiaohui Liang (\Letter) \at State Key Lab of Virtual Reality Technology and Systems, Beihang University, Beijing, China \\
\email{\{craaaaazy, zhengyijian, liang\_xiaohui\}@buaa.edu.cn}
\and
Xinning Wang, Jianguo Li (\Letter) \at Dept. of Rheumatology and Immunology, Children's Hospital, Capital Institute of Pediatrics, Beijing, China \\
\email{\{xnwangcip, jianguo\_li6\}@hotmail.com}
\and
% Yijian Zheng \at State Key Lab of Virtual Reality Technology and Systems, Beihang University, Beijing, China \\
% \email{zhengyijian@buaa.edu.cn}
% \and
% Liyuan Wang \at Dept. of Psychological and Cognitive Sciences, Tsinghua University, Beijing, China \\
% \email{liyuanwang@tsinghua.edu.cn}
% \and
% Jianguo Li (\Letter) \at Dept. of Rheumatology and Immunology, Ch ildren's Hospital, Capital Institute of Pediatrics, Beijing, China \\
% \email{jianguo\_li6@hotmail.com}
% \and
% Xiaohui Liang (\Letter) \at State Key Lab of Virtual Reality Technology and Systems, Beihang University, Beijing, China \\
% \email{liang\_xiaohui@buaa.edu.cn}
% \and
Xiaohui Liang \at Zhongguancun Laboratory, Beijing, China
}

\date{Received: date / Accepted: date}

\maketitle

\abstract{
    Action Quality Assessment (AQA) aims to evaluate how well a person performs a movement, which is essential in applications such as sports scoring, skill assessment, and healthcare. However, unimodal approaches often struggle to capture subtle cues of movement quality in real-world settings. Although multi-modal inputs provide complementary information, existing methods still face two major challenges: heterogeneous modalities often lead to cross-modal misalignment and unstable fusion, and reliable multi-modal annotation is costly, resulting in limited dataset diversity. To address these challenges, we propose DualAlign, a two-stage multi-modal fusion framework with adaptive alignment. The framework first constructs a coherent visual representation by maximizing shared structural information across RGB video, optical flow, and skeleton modalities. Textual semantics are then incorporated after visual stabilization, allowing high-level descriptions to complement rather than distort the underlying visual manifold. To evaluate the framework under realistic multi-modal conditions, we introduce MM--JDM, a movement-quality assessment dataset integrating RGB videos, optical flow, skeleton sequences, and structured text. MM--JDM naturally exhibits modality noise, class imbalance, and label scarcity, making it a challenging benchmark for studying multi-modal fusion and alignment. Extensive experiments show that DualAlign improves average correlation on MM--JDM by 21.16\% over the state-of-the-art methods and achieves gains of 3.53\% and 5.95\% on the RG and Fis-V benchmarks, respectively. DualAlign also remains robust under missing-modality and label-scarce conditions. 
}


\section{Introduction}

Action Quality Assessment (AQA) aims to quantify the execution quality and correctness of human movements
\cite{han2025caflow,xu2022finediving,dong2024interpretable}.
It plays an essential role in sports analysis \cite{parmar2017learning,pan2019action,dong2026uila}, skill assessment \cite{doughty2018s,gao2023automatic}, and healthcare
\cite{zhou2023video,liu2021towards,li2024egoexo}, where consistent and objective evaluation is essential.
Most existing AQA systems predominantly rely on unimodal video input \cite{zhou2025phi,xu2024fineparser,xu2022finediving,ke2024two}. However, RGB videos cannot explicitly represent structural cues such as body pose configurations or high-level semantic descriptions of movement quality \cite{zhou2026comprehensive,yin2026decade}. 
For example, two rhythmic gymnastics movements may appear visually similar in RGB frames while differing substantially in posture stability, which can be more explicitly captured by skeleton representations and further clarified by semantic descriptions. 

Although incorporating multiple modalities has the potential to alleviate the limitations of unimodal AQA, our empirical study reveals that existing multi-modal approaches still confront two fundamental challenges.
In this paper, we use the term ``\emph{multi-modal}'' to denote heterogeneous cues describing complementary aspects of human movement, including both independent sensing modalities (e.g., textual descriptions) and representations derived from visual inputs (e.g., optical flow). \textbf{First}, the heterogeneous nature of different modalities often leads to representation discrepancies that hinder effective cross-modal interaction and destabilize multi-modal fusion. In practice, this issue can reduce the expected benefits of multi-modal integration. We even observe cases where unimodal methods approach or outperform their multi-modal counterparts. For instance, the unimodal PHI model \cite{zhou2025phi} surpasses recent multi-modal methods \cite{xu2025language} on the Ball action of the RG dataset (see \cref{tab:rg}), a trend that is also reflected in broader comparisons on other datasets (see \cref{tab:fisv}).
\textbf{Second}, multi-modal annotation is costly and resource-intensive, which often results in limited dataset scale and modest modality diversity (see \cref{tab:MM-JDM-cmp}). These constraints substantially hinder progress in developing and rigorously evaluating more effective multi-modal strategies for AQA.

\begin{figure}
    \centering
    \includegraphics[width=\linewidth]{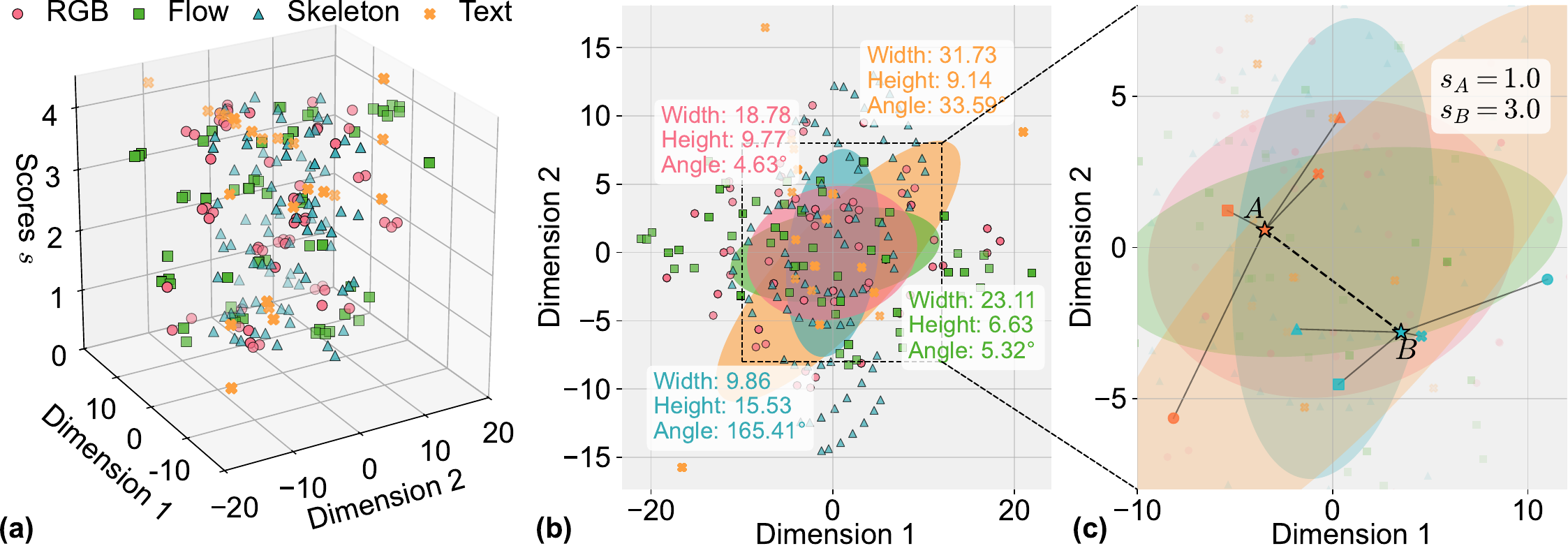}
    \caption{
        Illustration of cross-modal representation discrepancies across different inputs.
        We project MM--JDM features from different modalities into a low-dimensional space using t-SNE, with each modality denoted by a distinct marker.
        (a) shows the 3D embedding, where the $z$-axis corresponds to assessment scores.
        (b) presents the 2D projection, highlighting the intertwined and misaligned distributions across modalities.
        (c) visualizes two representative samples, where inconsistent cross-modal distributions illustrate the difficulty of learning a coherent shared representation.
    }
    {
        \phantomsubcaption\label{fig:key-challenge-a}
        \phantomsubcaption\label{fig:key-challenge-b}
        \phantomsubcaption\label{fig:key-challenge-c}
    }\label{fig:key-challenge}
\end{figure}

Movement-quality assessment in unconstrained environments further amplifies these challenges, as real-world human movements often exhibit substantial inter-subject variability, inconsistent execution quality, frequent self-occlusions, and limited annotated data. These factors complicate both reliable modeling and effective multi-modal fusion.
Juvenile Dermatomyositis (JDM), a pediatric condition associated with muscle weakness \cite{rider2018update,mccann2022juvenile}, provides a representative scenario in which these difficulties become particularly pronounced. Muscle Strength Assessment (MSA) for JDM typically relies on expert-scored protocols such as CMAS \cite{rider2018update}, which are subjective, labor-intensive, and difficult to scale. Although automated AQA offers a promising alternative, unimodal video alone \cite{zhou2023video} cannot capture the structural and semantic cues required to assess complex movement quality.
As illustrated in \cref{fig:key-challenge}, different modalities often exhibit heterogeneous feature distributions due to their distinct statistical properties and abstraction levels. Such cross-modal discrepancies hinder effective interaction between modalities, leading to unstable fusion and underutilization of modality-specific cues \cite{li2026multimodal}. Although end-to-end networks can theoretically learn fusion strategies automatically, directly combining such heterogeneous representations often results in suboptimal modality interaction, where dominant modalities suppress complementary cues. Consequently, existing multi-modal AQA methods suffer noticeable performance degradation when applied under these conditions (see \cref{tab:exp}). These observations position JDM-MSA as a representative and challenging testbed for developing principled multi-modal alignment strategies that can generalize to broader movement-quality assessment problems.

To address the challenge of cross-modal representation discrepancies, we propose \textbf{DualAlign}, a two-stage framework with adaptive alignment.
Our framework takes four complementary inputs describing human movement, including RGB video, optical flow, and skeleton representations derived from visual inputs, and textual descriptions, capturing appearance, motion, structural pose, and semantic information, respectively. The design is modality-agnostic and can be naturally extended to other sensing inputs.

The motivation for the two-stage design stems from the inherent disparity between visual and textual representations (see \cref{fig:key-challenge}). Visual modalities share relatively consistent spatiotemporal structures, whereas textual descriptions reside in a higher-level semantic space that is not directly compatible with early-stage visual feature formation. When textual features are fused into the visual stream too early, heterogeneous representations are prematurely forced into a shared embedding space, which can suppress modality-specific cues and destabilize the learning process. This phenomenon is empirically supported by the performance degradation observed in both the one-stage alignment variant and the reversed fusion order (see \cref{fig:stage}).
Accordingly, DualAlign first aligns visual modalities to establish a coherent and redundancy-reduced visual representation by maximizing shared structural information across RGB video, optical flow, and skeleton sequences. After the visual representation becomes stable, textual semantics are introduced to enrich high-level interpretation while preserving the integrity of the underlying visual patterns.

Nevertheless, fusion architectures alone remain insufficient to resolve cross-modal representation discrepancies (see \cref{tab:ablation}). Traditional approaches typically learn alignment through pairwise cross-modal similarities \cite{li2026multimodal}, which makes it difficult to capture consistent spatiotemporal relationships across more than two modalities simultaneously. GRAM \cite{cicchetti2024gramian} provides a geometric formulation that models the joint relationships among multiple modality vectors by minimizing the volume they span. However, the original GRAM formulation performs alignment across all modalities in a single stage, which is not well-suited for scenarios where modalities interact at different stages of representation learning. To address this limitation, we adapt the Gramian alignment principle to our two-stage architecture, enabling progressive alignment across modalities while preserving modality-specific information. Empirically, this design improves robustness under heterogeneous and incomplete multi-modal settings (see \cref{fig:miss-label-plot,fig:miss-modal-plot}).

\begin{table}[!t]
    \centering
    \caption{
    Comparison of open-source multi-modal AQA datasets. The available modalities include Video (V), Optical Flow (F), Skeleton (S), and Text (T).
    }
    \setlength{\tabcolsep}{6pt}
    \resizebox{\linewidth}{!}{
        \begin{tabular}{rccccm{1.0cm}<\centering m{1.0cm}<\centering m{2cm}<\centering}
            \toprule
            \multirow{2.5}{*}{\textbf{Dataset}}     
            & \multicolumn{4}{c}{\textbf{Modality}} & \multirow{2.5}{*}{\textbf{Classes}} & \multirow{2.5}{*}{\textbf{Samples}} & \multirow{2.5}{*}{\textbf{Annotation}} \\
            \cmidrule(lr){2-5}
                                                                 & \textbf{V}                            & \textbf{F}                        & \textbf{S} & \textbf{T} &              &                                         \\
            \midrule
            \rowcolor{orange!8} FineFS~\cite{ji2023localization} & \ding{51}                             &                                   & \ding{51}  &            & \phantom{0}4 & 1167           & Subaction Class, Score \\
            \rowcolor{yellow!8} UI-PRMD~\cite{vakanski2018data}  &                                       &                                   & \ding{51}  &            & 10           & 1326           & Binary Class           \\
            \rowcolor{yellow!8} KIMORE~\cite{capecci2019kimore}  & \ding{51}                             &                                   & \ding{51}  &            & \phantom{0}5 & 1560           & Grade                  \\
            \rowcolor{yellow!8} EHE~\cite{bruce2021skeleton}     &                                       &                                   & \ding{51}  &            & \phantom{0}6 & \phantom{0}869 & Binary Class           \\
            \rowcolor{brown!8} MM--JDM (Ours)                     & \ding{51}                             & \ding{51}                         & \ding{51}  & \ding{51}  & 12           & 1639           & Action Class, Grade    \\
            \bottomrule
        \end{tabular}
    }
    \label{tab:MM-JDM-cmp}
\end{table}

To address the scarcity of high-quality multi-modal data for movement-quality assessment, we construct \textbf{MM--JDM}, a substantially expanded dataset collected over an additional two-year period, extending our earlier RGB-only work \cite{zhou2023video}.
MM--JDM integrates RGB videos, optical flow, and skeleton sequences, which provide complementary appearance, motion, and structural cues, together with structured textual information such as action descriptions and examination notes.
All recordings are acquired from real-world assessments, and the annotation process covers action labels, movement-quality scores, and textual descriptions.
This process is conducted under strict expert verification to ensure reliability and consistency.
Compared with existing open-source multi-modal AQA datasets, MM--JDM offers a uniquely rich combination of modalities and a more challenging data distribution (see \cref{tab:MM-JDM-cmp,fig:dataset}).
This increased difficulty is evidenced by the pronounced performance degradation of state-of-the-art methods on MM–JDM (see \cref{tab:exp}), highlighting its value as a challenging benchmark for multi-modal fusion and alignment.
Although DualAlign is designed to advance multi-modal AQA, the comprehensive annotations and diverse movement patterns in MM--JDM also enable broader investigations into movement characteristics with potential relevance to clinical analysis.

Experiments demonstrate that DualAlign improves the average correlation on MM--JDM by 21.16\% over the second-best method and achieves consistent gains on two public AQA benchmarks (3.53\% on RG and 5.95\% on Fis-V).
Our main contributions are threefold:
\begin{itemize}
    \item We propose \textbf{DualAlign}, a two-stage multi-modal alignment framework that explicitly separates visual--visual and visual--textual alignment, thereby reducing cross-modal redundancy while preserving modality-specific structure for action quality assessment.

    \item We introduce \textbf{MM--JDM}, a multi-modal AQA dataset that integrates multiple visual modalities with structured text, thereby providing a challenging benchmark for studying cross-modal alignment under realistic imbalance and label-scarce conditions.

    \item We conduct extensive experiments and analyses that demonstrate the effectiveness of staged alignment and the complementary roles of heterogeneous modalities across multiple AQA benchmarks, including robustness to missing modalities and limited supervision.
\end{itemize}

The remainder of this paper is organized as follows.
\cref{sec:related-work} reviews related work.
\cref{sec:method} describes the proposed DualAlign framework and its core components.
\cref{sec:dataset} introduces the MM--JDM dataset and outlines its construction and statistical properties.
\cref{sec:experiments} reports experimental results and analysis.
Finally, \cref{sec:conclusion} concludes the paper.

\section{Related Work} \label{sec:related-work}
This section reviews relevant literature on AQA methods and multi-modal alignment learning.

\subsection{Action Quality Assessment (AQA)}
AQA involves the quantitative evaluation of action performance and provides objective and consistent measures of movement quality \cite{zhou2026comprehensive}. Different from action recognition \cite{zhao2021classifying,zhao2022place}, which mainly concerns semantic action categorization, AQA emphasizes fine-grained performance evaluation by measuring the quality, proficiency, or correctness of an action execution. It plays an important role in a wide range of applications, including sports performance analysis \cite{li2022pairwise,yu2021group,zhou2023hierarchical,zhou2024cofinal,liu2023figure,xu2022likert,zhou2024magr,zhou2025continual}, rehabilitation monitoring \cite{deb2022graph,bruce2024egcn++}, and professional skill assessment \cite{bertasius2017baller,fang2024better,yan2026cel}.

Historically, AQA methods relied on a single modality of data to assess action quality. In sports analysis \cite{li2022pairwise,zhou2023hierarchical,han2025finecausal,xu2022likert,zhou2024magr,li2024continual}, playback videos of Olympic games are commonly used as inputs to evaluate athletes' actions. Its accessibility drives this reliance on video data. However, skeletal data is often underutilized in such scenarios due to the limitations of pose estimation algorithms, which struggle with noise and the complex motions characteristic of athletic performances \cite{wang2021tsa,pirsiavash2014assessing,xu2024procedure,xu2025human,xu2024fineparser,li2024egoexo}.
Conversely, in medical care \cite{zhou2023video,deb2022graph,bruce2024egcn++}, where precision is paramount, skeletal data becomes the primary modality for assessing patients' conditions. These applications demand high accuracy and often involve capturing skeletal data using depth cameras, which provide reliable inputs for monitoring motion and posture.
Recently, the integration of multi-modal data in AQA has gained prominence due to its capability to leverage diverse information sources for a more comprehensive assessment. Modalities such as video, audio, and skeletal data offer complementary perspectives that collectively enhance the accuracy, robustness, and contextual understanding of action quality. By combining visual, auditory, and kinematic cues, multi-modal AQA systems \cite{du2023learning,ji2023localization,xia2023skating,zeng2024multimodal,xu2024vision,xu2025quality} can capture subtle cues of actions that may be obscured or incomplete when analyzed using a single modality alone.
For example, Ji et al. \cite{ji2023localization} focus on integrating skeletal and visual cues to localize and evaluate precise movements in various action sequences and contribute a new multi-modal AQA dataset. Similarly, Xia et al. \cite{xia2023skating} leverage both audio and visual cues to assess the performance of figure skating.

However, existing multi-modal AQA methods often struggle with cross-modal representation discrepancies across different inputs, which limits their effectiveness in real-world scenarios. In applications such as JDM-MSA, the interaction between textual descriptions and visual data introduces additional complexity that requires more specialized alignment strategies. Our work addresses these challenges by proposing a method designed to improve the integration of heterogeneous inputs for multi-modal AQA.

\subsection{Multi-Modal Alignment}
Multi-modal integration enhances model accuracy and applicability by leveraging complementary information across diverse modalities.
A core challenge in multi-modal fusion lies in alignment, which aims to establish consistent semantic relationships across modalities so that their representations can be coherently projected into a shared latent space~\cite{li2026multimodal}.

Existing multi-modal alignment methods \cite{kruskal1983overview,hotelling1992relations,radford2021learning,zhang2022pointclip,luo2022clip4clip} can be categorized into explicit and implicit approaches.
Among implicit alignment approaches, CLIP~\cite{radford2021learning} has been particularly influential, establishing a scalable paradigm for aligning image and text representations. This paradigm has since been extended to other modality pairs, including audio--text (CLAP~\cite{elizalde2023clap}), video--text (CLIP4Clip~\cite{luo2022clip4clip}), and point cloud--text (PointCLIP~\cite{zhang2022pointclip}).
More recent extensions further incorporate additional modalities through anchor-based alignment strategies, such as CLIP4VLA~\cite{ruan2023accommodating}.
However, these approaches typically align each modality independently to a designated anchor without explicitly enforcing consistency across all modalities, which can limit their effectiveness in tasks that require holistic multi-modal understanding~\cite{girdhar2023imagebind,zhu2024languagebind,zhao2024videoprism}.
To address this limitation, GRAM~\cite{cicchetti2024gramian} was recently proposed to align more than two modalities simultaneously by minimizing geometric volume in a shared embedding space, achieving strong performance on large-scale retrieval and classification benchmarks.
In parallel, large multi-modal models (LMMs), such as SEED-X~\cite{cheng2025seed} and Qwen2-VL~\cite{bai2025qwen2}, attempt to resolve semantic heterogeneity through massive model scaling and unified architectures.
Despite their impressive generalization capabilities, directly applying these heavy-weight models to AQA remains challenging.
First, AQA datasets in domains such as sports and medical assessment are typically small and label-scarce~\cite{zhou2026comprehensive}, which makes billion-parameter models prone to overfitting and unstable optimization.
Second, although GRAM provides a principled foundation for multi-party alignment, its simultaneous fusion strategy does not explicitly account for the heterogeneous structure of visual and textual modalities. As a result, introducing textual semantics at an early stage can suppress action-centric visual cues, a limitation that is empirically confirmed by the performance degradation observed in~\cref{tab:exp}.

These limitations motivate the design of a lightweight and staged alignment framework that preserves visual structure while still benefiting from complementary semantic information.
Unlike vanilla GRAM, which aligns all modalities simultaneously, we propose a two-stage alignment strategy.
The first stage aligns only visual modalities to form a coherent spatiotemporal representation, followed by a second stage that introduces textual information to complement the visual manifold.
This staged design explicitly respects modality heterogeneity and enables more effective multi-modal integration.

\section{DualAlign: Two-Stage Aligned Fusion} \label{sec:method}
This section first provides an overview of the DualAlign framework and then details its core components.
In the following, we use \emph{alignment} to describe cross-modal representation matching and \emph{fusion} to denote the integration of modality features.

\subsection{Motivation and Method Overview}
\label{sec:motivation}

\myPara{Choice of Inputs}
In this work, we consider four complementary inputs for human movement analysis: RGB video, optical flow, skeleton data, and text. These inputs capture different aspects of movement quality, including appearance, motion dynamics, structural pose, and high-level semantic descriptions. Importantly, all these modalities are readily available in common video-based analysis pipelines \cite{zeng2024multimodal,xu2025language,xu2025quality}. Optical flow and skeleton representations can be reliably extracted from RGB videos using standard motion and pose estimation tools, while textual descriptions can be obtained from existing annotations or expert comments in many AQA datasets. This makes the proposed framework practical for real-world scenarios where multiple sources of information may be available.

\myPara{Challenges}
Despite their complementary nature, heterogeneous modalities often exhibit distinct statistical properties and semantic abstractions, which introduce cross-modal representation discrepancies (see \cref{fig:key-challenge}). When modalities with such discrepancies are fused directly, their incompatible representations may hinder effective cross-modal interaction and lead to unstable multi-modal fusion.
Existing approaches \cite{xu2024vision,xia2023skating,du2023learning} typically rely on standard fusion architectures that implicitly assume modality compatibility. In practice, however, this assumption rarely holds, and naïve multi-modal fusion may fail to fully exploit complementary information.
To move beyond unimodal analysis, we extend our previous work \cite{zhou2023video} by incorporating additional modalities for a more comprehensive assessment of movement quality. However, empirical results consistently show that directly integrating these modalities may even degrade performance. This is observed both in prior benchmarks (see \cref{tab:rg,tab:fisv}) and in our multi-modal JDM-MSA experiments (see \cref{tab:exp}), indicating that naïve multi-modal integration alone is insufficient to resolve the representation discrepancies.

\myPara{Motivation}
To address these representation discrepancies, a key insight is that heterogeneous modalities should be aligned according to their structural properties rather than fused directly. Traditional pairwise alignment strategies \cite{li2026multimodal} struggle to model consistent relationships among multiple modalities, limiting their ability to capture coherent spatio-temporal cues across more than two inputs.
GRAM \cite{cicchetti2024gramian} provides a geometric formulation that enables joint alignment across multiple modalities. However, its single-stage formulation does not account for the structural differences between visual and semantic representations, which may lead to unstable multi-modal integration (see \cref{tab:ablation}).
These observations suggest that alignment should be performed progressively rather than jointly. In particular, visual modalities share relatively homogeneous spatio-temporal structures and can therefore be aligned first to establish a coherent action-centered representation. Textual semantics lie in a fundamentally different feature space and are introduced only after the visual representation stabilizes, allowing high-level descriptions to enrich rather than distort the visual embedding. 

\begin{figure*}
    \centering
    \begin{overpic}[width=\linewidth,clip,trim=10 75 15 80]{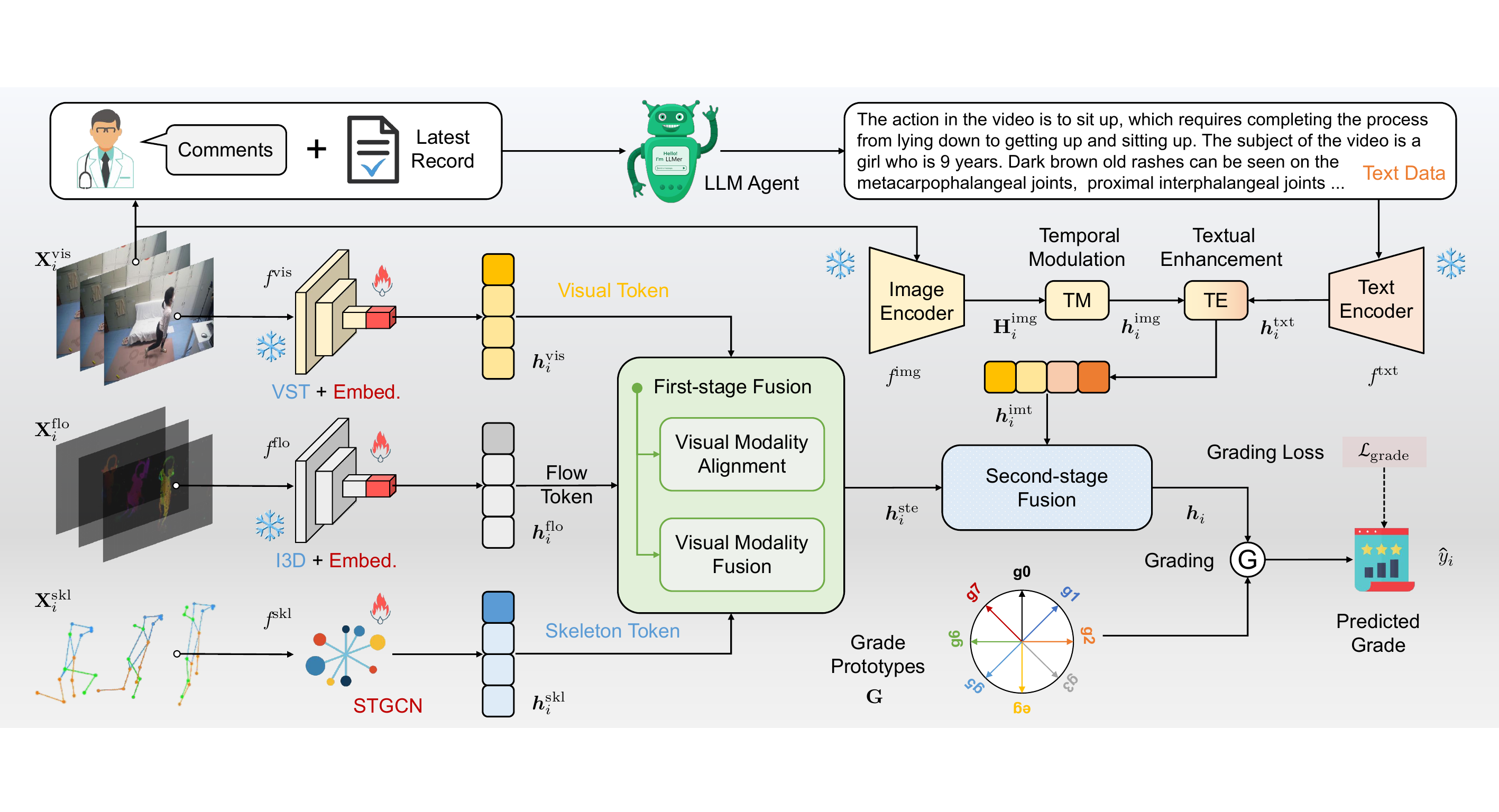}

    \end{overpic}
    \caption{
        The core insight of DualAlign is leveraging a two-stage fusion design with adaptive alignment to fully exploit the complementarity of diverse inputs, thereby improving action assessment performance.
        In the first stage, we align and fuse video, flow, and skeleton data. Flow and skeleton provide vital temporal and spatial cues for accurate assessment, enhancing the visual data representation. In the second stage, the enhanced visual data is aligned and merged with textual data to produce the final representation.
        Finally, the final representation is used to predict the grades by using pre-defined grade prototypes.
        \includegraphics[height=\myMheight]{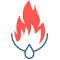} and \includegraphics[height=\myMheight]{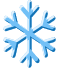} indicate the learnable and fixed statuses of pre-trained backbones, respectively.
    }
    \label{fig:framework}
\end{figure*}

\myPara{Notation}
In the following, the subscript $i$ in each notation denotes the $i$-th sample in the dataset.
Each sample includes two modality groups: visual data and textual data.
The visual data consists of video data $\mathbf{X}_i^{\mathrm{vis}} \in \mathbb{R}^{T \times W \times H \times 3}$, flow data $\mathbf{X}_i^{\mathrm{flo}} \in \mathbb{R}^{T \times W \times H \times 2}$, and skeleton data $\mathbf{X}^{\mathrm{skl}}_i \in \mathbb{R}^{T \times J \times 3}$, where $T$, $W$, $H$, and $J$ denote the number of frames, width, height, and the number of joints, respectively. These modalities provide complementary temporal and spatial information, which is critical for accurate assessment. To effectively capture the underlying features of each modality, we employ modality-specific encoders $f^{\mathrm{vis}}$, $f^{\mathrm{flo}}$, and $f^{\mathrm{skl}}$ to extract representations $\bm{h}^{\mathrm{vis}}_i \in \mathbb{R}^{D_1}$, $\bm{h}^{\mathrm{flo}}_i \in \mathbb{R}^{D_1}$, and $\bm{h}^{\mathrm{skl}}_i \in \mathbb{R}^{D_1}$ from $\mathbf{X}_i^{\mathrm{vis}}$, $\mathbf{X}_i^{\mathrm{flo}}$, and $\mathbf{X}_i^{\mathrm{skl}}$. Here, we will use $D$ with distinct subscripts, such as $D_1$, $D_2$, etc., to differentiate between feature dimensions.
To prevent overfitting caused by the small sample size per action, we freeze the vanilla backbones for video and flow data. Additionally, lightweight embedding layers are appended to these backbones to enhance adaptability (see \cref{fig:framework}).
The textual data process can be seen in \cref{sec:dataset_construction}.
Similar to \citet{xu2024vision}, we employed a pre-trained vision-language model \cite{radford2021learning} to obtain the textual embedding $\bm{h}_i^{\mathrm{txt}} \in \mathbb{R}^{D_2}$. 

\myPara{Framework Overview}
To address cross-modal representation discrepancies, we design DualAlign, a two-stage alignment framework for multi-modal action assessment. The overall architecture is illustrated in \cref{fig:framework}.
In the first stage (see \cref{sec:first}), we aim to align and fuse the visual modalities into an enhanced representation $\bm{h}_i^{\mathrm{ste}} \in \mathbb{R}^{D_2}$.
In the second stage (see \cref{sec:second}), the enhanced visual representation $\bm{h}_i^{\mathrm{ste}}$ is aligned with the textual data $\bm{h}_i^{\mathrm{txt}}$ to produce a final joint representation $\bm{h}_i$.
Finally, the fused representation $\bm{h}_i \in \mathbb{R}^{D_3}$ is used to predict the action score $\hat{y}_i$ via a prototype-based prediction mechanism. This strategy has been shown in our previous work \cite{zhou2024cofinal} to improve robustness and accuracy in small-scale AQA settings, and we follow the same configuration for continuous regression.
For discrete assessment tasks, the score is defined on a grading scale with $G$ levels, where $G$ denotes the number of grade categories for a given action (see \cref{tab:MM-JDM-stat} for MM--JDM). Each grade is represented by a prototype vector in the set $\mathbf{G} = [\bm{g}_1, \bm{g}_2, \cdots, \bm{g}_G]$, constructed using a simplex Equiangular Tight Frame (ETF) structure \cite{kothapalli2023neural,zhu2021geometric}, where each prototype corresponds to one grade level and prototypes are mutually perpendicular to each other in the embedding space.  The predicted grade is obtained by selecting the closest prototype:
\begin{equation} \label{eq:grade}
    \hat{y}_i = \arg\min_{j} \| \bm{h}_i - \bm{g}_j \|_2, \quad j = 1, 2, \cdots, G.
\end{equation}
For continuous regression tasks, directly increasing the number of prototypes to achieve higher prediction resolution may lead to inefficiency and unstable optimization. To address this, we adopt a coarse-to-fine prediction strategy, following \citet{zhou2024cofinal}, which first estimates a coarse-grained grade and a fine-grained grade (using a similar formulation as \cref{eq:grade}). The final prediction is obtained by combining these two levels.

\myPara{Problem Formulation}
Following the prototype-based prediction mechanism in \cref{eq:grade}, we train the model by encouraging the learned representation $\bm{h}_i$ to be close to the prototype corresponding to its ground-truth grade. Accordingly, we adopt a classification loss for grade prediction:
\begin{equation}
    \mathcal{L}_{\mathrm{grade}} = - \frac{1}{N} \sum_{i=1}^N \sum_{j=1}^G p_{ij} \log\left(\frac{\exp(\bm{g}_j \cdot \bm{h}_i)}{\sum_{k=1}^{G} \exp(\bm{g}_k \cdot \bm{h}_i)}\right),
\end{equation}
where $N$ is the number of samples and $p_{ij}$ denotes the one-hot label of the ground-truth grade $y_i$. 
Both the prototypes $\bm{g}_j$ and the representation $\bm{h}_i$ are normalized to unit length, which makes maximizing the dot-product similarity equivalent to minimizing the Euclidean distance in \cref{eq:grade}. However, this prediction-oriented loss alone does not explicitly address cross-modal representation discrepancies. To this end, DualAlign introduces additional alignment objectives to regularize the feature space. The overall objective is formulated as:
\begin{equation}
    \begin{aligned}
        \min_{\Theta}~ & \mathcal{L} = \mathcal{L}_{\mathrm{grade}} + \lambda_1 \mathcal{L}_{\mathrm{vfs}} + \lambda_2 \mathcal{L}_{\mathrm{vit}},
    \end{aligned}
\end{equation}
where $\Theta$ denotes the learnable parameters, $\lambda_1$ and $\lambda_2$ are trade-off coefficients, and $\mathcal{L}_{\mathrm{vfs}}$ (see \cref{eq:vfs}) and $\mathcal{L}_{\mathrm{vit}}$ (see \cref{eq:vit}) denote the alignment losses for visual modalities and visual-text modalities, respectively.

\subsection{First-Stage Fusion} \label{sec:first}

\begin{figure}
    \centering
    \begin{overpic}[width=\linewidth,clip,trim=280 200 280 205]{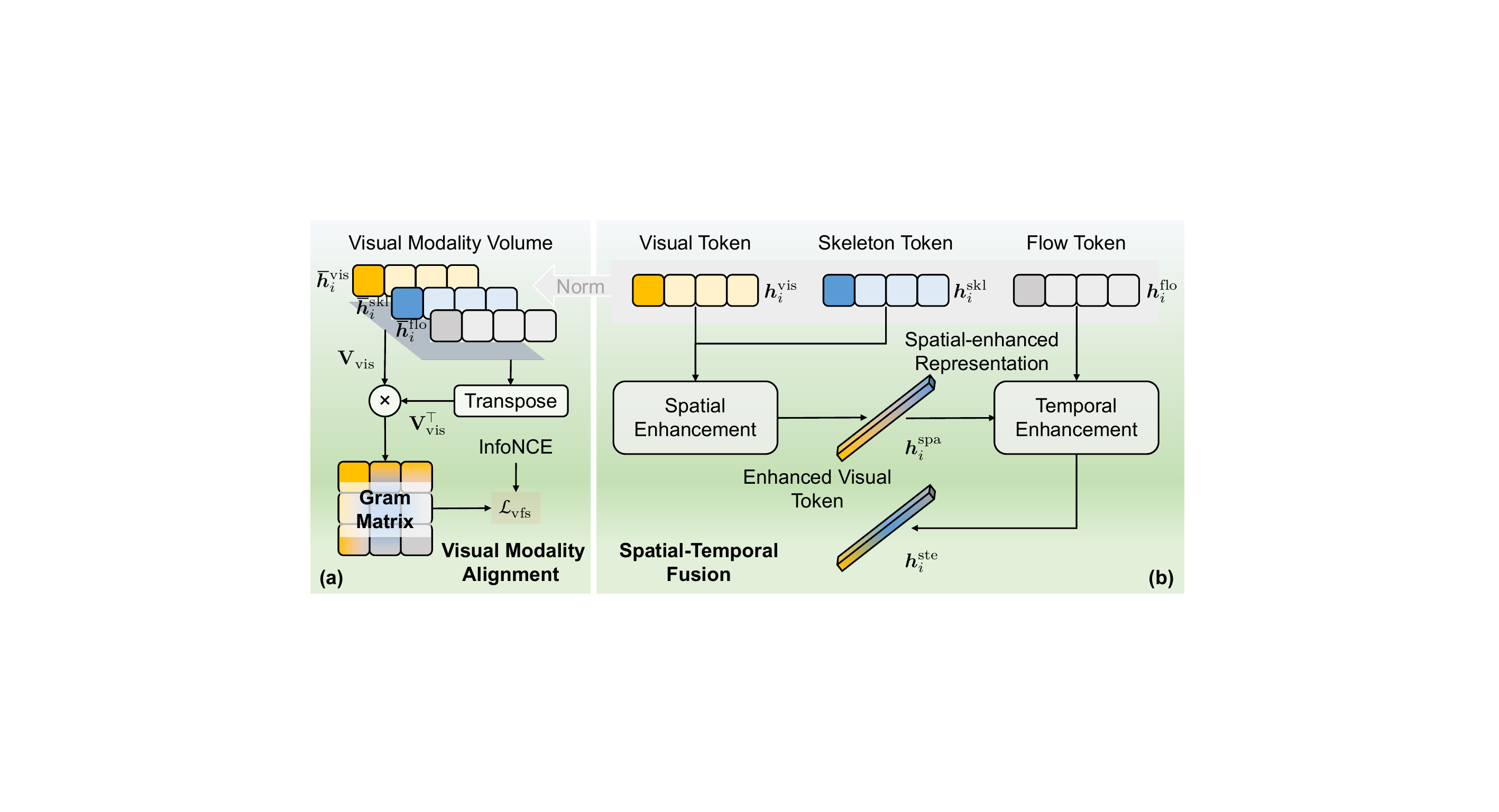}
    \end{overpic}
    \caption{Illustrations of the first-stage fusion process: (a) depicts the alignment of visual modalities (video, flow, and skeleton), ensuring a shared representation space, and (b) illustrates the enhancement of visual data through the integration of complementary spatial and temporal cues.}
    \label{fig:first-stage}{
        \phantomsubcaption\label{fig:first-stage-a}
        \phantomsubcaption\label{fig:first-stage-b}
    }
\end{figure}

\myPara{Justification}
The first-stage fusion is crucial for enhancing visual data representation by leveraging the complementary strengths of video, flow, and skeleton modalities while eliminating the redundancy inherent in video data. Video data provides a rich visual context, flow data captures critical motion dynamics, and skeleton data encodes spatial and structural cues. Relying solely on video data can be inefficient and may fail to capture the full complexity of actions. The first-stage fusion (see \cref{fig:first-stage}) constructs a robust visual representation by aligning (see \cref{fig:first-stage-a}) and merging (see \cref{fig:first-stage-b}) these modalities. This approach ensures the maximization of spatial and temporal information, which is particularly critical for accurately assessing complex actions where individual modalities may omit essential details.

\myPara{Visual Modality Alignment}
While modality-specific encoders retain modality-specific features, their isolated optimization leads to suboptimal performance due to the misalignment issue.
To achieve a unified representation across visual modalities, we address the challenges of misalignment using a combination of a Gram matrix-based alignment and a novel multi-modal contrastive loss function.

\begin{definition}[GRAM Multi-Modal Alignment Loss \cite{cicchetti2024gramian}] \label{def_mm_loss}
    For $K$ visual modality vectors $\{\bm{h}_i^1, \bm{h}_i^2, \cdots, \bm{h}_i^K\}$, we normalize each to obtain $\{\bar{\bm{h}}_i^1, \bar{\bm{h}}_i^2, \cdots, \bar{\bm{h}}_i^K\}$. A designated anchor modality $\bm{a}_i$ serves as the central reference for alignment, while the remaining modalities are denoted as $\{\bm{m}_i^2, \bm{m}_i^3, \cdots, \bm{m}_i^K\}$. The GRAM multi-modal alignment loss is defined as a combination of two components:
    \begin{equation}
        \mathcal{L}_{\mathrm{gram}} = \mathcal{L}_{\mathrm{da}} + \mathcal{L}_{\mathrm{ad}},
    \end{equation}
    where the direct-anchor loss $\mathcal{L}_{\mathrm{da}}$ aligns all modalities to the anchor, and the anchor-direct loss $\mathcal{L}_{\mathrm{ad}}$ ensures bidirectional consistency. These are formulated as:
    \begin{equation}
        \mathcal{L}_{\mathrm{da}} = -\frac{1}{B} \sum_{i=1}^B \log \frac{\exp \left(-\operatorname{Vol}\left(\bm{a}_i, \bm{m}^{2}_i, \cdots, \bm{m}^{K}_i\right) / \tau\right)}{\sum_{j=1}^K \exp \left(-\operatorname{Vol}\left(\bm{a}_j, \bm{m}^{2}_i, \cdots, \bm{m}^{K}_i\right) / \tau\right)},
    \end{equation}
    \begin{equation}
        \mathcal{L}_{\mathrm{ad}} = -\frac{1}{B} \sum_{i=1}^B \log \frac{\exp \left(-\operatorname{Vol}\left(\bm{a}_i, \bm{m}^{2}_i, \cdots, \bm{m}^{K}_i\right) / \tau\right)}{\sum_{j=1}^K \exp \left(-\operatorname{Vol}\left(\bm{a}_i, \bm{m}^{2}_j, \cdots, \bm{m}^{K}_j\right) / \tau\right)},
    \end{equation}
    where $B$ is the batch size, $\tau$ is the temperature parameter, and $\operatorname{Vol}(\cdot)$ denotes the Gramian volume calculated using the determinant of the Gram matrix,
    which can be computed as:
    \begin{equation}
        \operatorname{Vol}\left(\mathbf{v}_1, \mathbf{v}_2, \cdots, \mathbf{v}_K\right) = \sqrt{\operatorname{det} \mathbf{M}\left(\mathbf{v}_1, \mathbf{v}_2, \cdots, \mathbf{v}_K\right)},
    \end{equation}
    where the Gram matrix $\mathbf{M}\in \mathbb{R}^{K \times K}$ is defined as:
    \begin{equation}
        \mathbf{M}\left(\mathbf{v}_1, \mathbf{v}_2, \cdots, \mathbf{v}_K\right) = \mathbf{V}^{\top} \mathbf{V},
    \end{equation}
    and $\mathbf{V}$ is the stacked matrix of modality vectors $\{\mathbf{v}_1, \mathbf{v}_2, \cdots, \mathbf{v}_K\}$. Each entry of $\mathbf{M}$ is computed as:
    \begin{equation}
        \mathbf{M}_{ij} = \langle \mathbf{v}_i, \mathbf{v}_j \rangle,
    \end{equation}
    where $\langle \cdot, \cdot \rangle$ denotes the inner product. The Gramian operation ensures that the volume spanned by the modality vectors is minimized, aligning their features in a shared latent space.
\end{definition}

In our scenario, we extract embedding vectors $\bm{h}_i^{\mathrm{vis}}$, $\bm{h}_i^{\mathrm{flo}}$, and $\bm{h}_i^{\mathrm{skl}}$ using modality-specific encoders for video, flow, and skeleton data, respectively. These vectors are then normalized to obtain unitary representations $\bar{\bm{h}}_i^{\mathrm{vis}}$, $\bar{\bm{h}}_i^{\mathrm{flo}}$, and $\bar{\bm{h}}_i^{\mathrm{skl}}$. Among these modalities, we designate the video data as the anchor modality, leveraging its inherent advantages as a comprehensive visual modality.
Video data encompasses rich spatiotemporal information, which serves as a natural bridge between spatial features from skeleton data and temporal motion cues from optical flow. By selecting video as the anchor, we align the skeleton and flow modalities with a robust representation, ensuring that all modalities converge toward a unified latent space with a high degree of intermodality coherence.

We reformulate the alignment loss (\cref{def_mm_loss}) as the visual-flow-skeleton alignment loss, formulated by:
\begin{equation} \label{eq:vfs}
    \begin{aligned}
        \mathcal{L}_{\mathrm{vfs}} = &
        -\frac{1}{B} \sum_{i=1}^B \log \frac{\exp \left(-\operatorname{Vol}\left(\bm{a}_i, \bar{\bm{h}}_i^{\mathrm{flo}}, \bar{\bm{h}}_i^{\mathrm{skl}}\right) / \tau\right)}{\sum_{j=1}^K \exp \left(-\operatorname{Vol}\left(\bm{a}_j, \bar{\bm{h}}_i^{\mathrm{flo}}, \bar{\bm{h}}_i^{\mathrm{skl}}\right) / \tau\right)}                                \\
                                     & -\frac{1}{B} \sum_{i=1}^B \log \frac{\exp \left(-\operatorname{Vol}\left(\bm{a}_i, \bar{\bm{h}}_i^{\mathrm{flo}}, \bar{\bm{h}}_i^{\mathrm{skl}}\right) / \tau\right)}{\sum_{j=1}^K \exp \left(-\operatorname{Vol}\left(\bm{a}_i, \bar{\bm{h}}_j^{\mathrm{flo}}, \bar{\bm{h}}_j^{\mathrm{skl}}\right) / \tau\right)}
    \end{aligned},
\end{equation}
where video is chosen as the anchor modality $\bm{a}_i = \bar{\bm{h}}_i^{\mathrm{vis}}$ due to its comprehensive representation of spatial and temporal cues. Unlike flow or skeleton features, video provides richer context, ensuring effective alignment of all modalities in a shared latent space.
This ensures that all modalities align toward a shared latent space for effective fusion.

\myPara{Visual Modality Fusion}
With modality alignment imposed during training to improve cross-modal consistency, the model integrates spatial and temporal cues to construct an enhanced visual representation. This design is essential for capturing both spatial structures and temporal dynamics in human motion.

Skeleton data encodes spatial and structural information, which is essential for understanding poses and movements. To leverage this information, we fuse skeleton features with video features using a spatial enhancement function:
\begin{equation}
    \bm{h}_i^{\mathrm{spa}} = \mathrm{SpatialEn}(\bm{h}_i^{\mathrm{vis}}, \bm{h}_i^{\mathrm{skl}}),
\end{equation}
where $\mathrm{SpatialEn}(\cdot)$ denotes the spatial enhancement function designed to effectively integrate the complementary information from the two modalities.

Flow data, on the other hand, captures motion dynamics and provides vital temporal information about changes in the scene. After spatial integration, temporal cues are incorporated through a temporal enhancement function:
\begin{equation}
    \bm{h}_i^{\mathrm{ste}} = \mathrm{TemporalEn}(\bm{h}_i^{\mathrm{spa}}, \bm{h}_i^{\mathrm{flo}}),
\end{equation}
where $\mathrm{TemporalEn}(\cdot)$ combines the spatially enhanced features with the temporal features derived from the flow data.

By systematically integrating spatial and temporal cues, this approach ensures that the visual representation captures the essential features necessary for downstream tasks.
The resultant representation, enriched with both spatial and temporal cues, is subsequently passed to a second-stage fusion network for further alignment with textual modalities. Notably, the spatial and temporal enhancement functions, $\mathrm{SpatialEn}(\cdot)$ and $\mathrm{TemporalEn}(\cdot)$, can adopt various architectures, such as Multi-Layer Perceptrons (MLPs) or cross-attention mechanisms, to optimize the fusion effectively.

\subsection{Second-Stage Fusion} \label{sec:second}

\begin{figure}
    \centering
    \begin{overpic}[width=\linewidth,clip,trim=280 200 280 205]{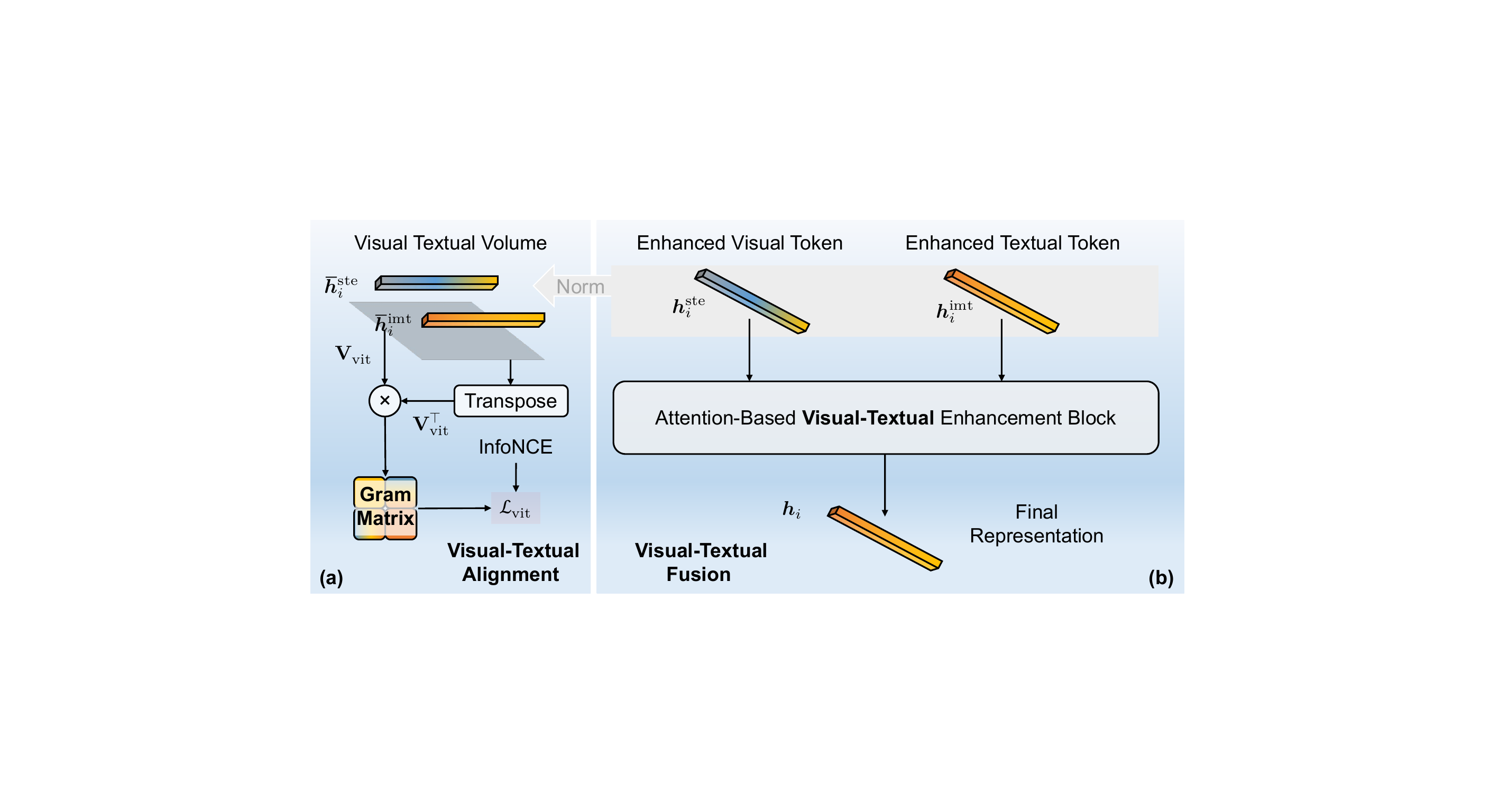}
    \end{overpic}
    \caption{Illustrations of the second-stage fusion process: (a) depicts the visual-textual alignment step, where embeddings from both modalities are aligned into a shared latent space; (b) shows the visual-textual fusion step, where aligned embeddings are integrated to form a unified representation.}
    \label{fig:second-stage}{
        \phantomsubcaption\label{fig:second-stage-a}
        \phantomsubcaption\label{fig:second-stage-b}
    }
\end{figure}

\myPara{Justification}
The second-stage fusion integrates textual modalities with enhanced visual representation, providing semantic context critical for domain-specific tasks. Textual data, such as expert comments or medical records, complements visual features by encoding semantic details that are not visually discernible. 
To achieve this, we first extract enhanced textual embeddings (see \cref{fig:framework}) and then align (see \cref{fig:second-stage-a}) and integrate (see \cref{fig:second-stage-b}) the textual and visual embeddings into a final representation. This facilitates robust performance in complex application scenarios. 
Textual input is optional at inference time and can be replaced with an empty placeholder when unavailable. The model remains effective using visual inputs alone, while training with text helps improve representation learning, as validated by our missing-modality experiments (see \cref{fig:miss-modal-plot}). 
\myPara{Initial Vision-Language Fusion}
Following \citet{xu2024vision}, we leverage the knowledge from both the pre-trained image encoder and text encoder \cite{radford2021learning}.
Although the enhanced visual token is already obtained from the first-stage fusion, CLIP's image embeddings offer semantic richness and a shared latent space with textual embeddings. In addition, CLIP's pre-training on large-scale vision-language data ensures robust generalization and enhances the textual modality with visual semantics beyond raw spatial or temporal features.

We process image sequences $\mathbf{X}^{\mathrm{vis}}_i$ through the pre-trained image encoder $f^{\mathrm{img}}$ to extract the visual token $\mathbf{H}^{\mathrm{img}}_i \in \mathbb{R}^{T\times D_2}$.  The visual token is complemented by textual token $\bm{h}_i^{\mathrm{txt}} \in \mathbb{R}^{D_3}$ obtained via the pre-trained text encoder $f^{\mathrm{txt}}$.
To address the limitations of the image encoder in capturing temporal dynamics, we go beyond the simplistic temporal pooling approach \cite{xu2024vision} by introducing a more advanced temporal modulation block. This block results in a richer representation:
\begin{equation} \bm{h}_i^{\mathrm{img}} = \mathrm{TemporalModulate}(\mathbf{H}^{\mathrm{img}}_i), \end{equation} where $\mathrm{TemporalModulate}(\cdot)$ preserves temporal coherence and captures subtle temporal dependencies across the image sequence.
Then, we enhance the textual token $\bm{h}_i^{\mathrm{txt}}$ using the temporally modulated visual features, producing the initial enhanced textual token:
\begin{equation}
    \bm{h}_i^{\mathrm{imt}} = \mathrm{EnhanceText}(\bm{h}_i^{\mathrm{txt}}, \bm{h}_i^{\mathrm{img}}),
\end{equation}
where $\mathrm{EnhanceText}(\cdot)$ integrates the visual and textual information through a fusion mechanism, such as cross-attention or a feedforward neural network.

\myPara{Visual-Textual Alignment}
Despite the enhanced textual embedding $\bm{h}_i^{\mathrm{imt}}$ and the enhanced visual embedding $\bm{h}_i^{\mathrm{ste}}$ from the first-stage fusion, the misalignment issue persists. Direct fusion of these embeddings may degrade performance. Leveraging the insights from \cref{def_mm_loss}, we first normalize the tokens $\bm{h}_i^{\mathrm{ste}}$ and $\bm{h}_i^{\mathrm{imt}}$ to get $\bar{\bm{h}}_i^{\mathrm{ste}}$ and $\bar{\bm{h}}_i^{\mathrm{imt}}$.
The visual-textual alignment loss $\mathcal{L}_{\mathrm{vit}}$ is formulated as:
\begin{equation} \label{eq:vit}
    \begin{aligned}
        \mathcal{L}_{\mathrm{vit}} =& -\frac{1}{B} \sum_{i=1}^B \log \frac{\exp \left(-\operatorname{Vol}\left(\bar{\bm{h}}_i^{\mathrm{ste}}, \bar{\bm{h}}_i^{\mathrm{imt}}\right) / \tau\right)}{\sum_{j=1}^K \exp \left(-\operatorname{Vol}\left(\bar{\bm{h}}_j^{\mathrm{ste}}, \bar{\bm{h}}_i^{\mathrm{imt}}\right) / \tau\right)} \\ &-\frac{1}{B} \sum_{i=1}^B \log \frac{\exp \left(-\operatorname{Vol}\left(\bar{\bm{h}}_i^{\mathrm{ste}}, \bar{\bm{h}}_i^{\mathrm{imt}}\right) / \tau\right)}{\sum_{j=1}^K \exp \left(-\operatorname{Vol}\left(\bar{\bm{h}}_i^{\mathrm{ste}}, \bar{\bm{h}}_j^{\mathrm{imt}}\right) / \tau\right)}. \end{aligned}
\end{equation}
Here, the enhanced visual modality  $\bar{\bm{h}}_i^{\mathrm{ste}}$ is chosen as the anchor because spatial-temporal enhancement directly integrates spatial structure and temporal motion, capturing comprehensive information critical for alignment. In contrast, the enhanced textual modality, derived from CLIP, primarily augments semantic context without fully leveraging temporal coherence. Thus, $\bar{\bm{h}}_i^{\mathrm{ste}}$ provides a stronger foundation for aligning diverse modalities. This alignment step ensures visual and textual modalities converge toward a shared latent space for effective fusion.

\myPara{Visual-Textual Fusion}
Finally, we enhance the aligned visual and textual embeddings by integrating them using a cross-modal enhancement function:
\begin{equation}
    \bm{h}_i = \mathrm{Enhance}({\bm{h}}_i^{\mathrm{ste}}, {\bm{h}}_i^{\mathrm{imt}}), \end{equation}
where $\mathrm{Enhance}(\cdot)$ is a function such as a cross-attention module or an MLP that effectively combines the modalities. The unified embedding $\bm{h}_i$ captures complementary visual and textual semantics, enabling accurate action assessment.

\section{MM-JDM: The Multi-Modal JDM Dataset} \label{sec:dataset}
This section presents the motivation behind developing the MM--JDM dataset, outlines the dataset construction process, and provides an overview of its statistical characteristics.

\subsection{Dataset Motivation}
As summarized in \cref{tab:MM-JDM-cmp}, existing multi-modal AQA datasets offer limited modality diversity, annotation depth, and real-world variability, which constrains systematic investigation of multi-modal fusion and alignment. In pediatric movement-quality assessment, recordings frequently exhibit substantial variability, occlusions, and inconsistent execution patterns. Relying solely on RGB video \cite{zhou2023video} cannot adequately capture the structural or semantic cues required for reliable assessment, a limitation that is reflected in the performance drop observed in \cref{tab:ablation}. To overcome these shortcomings, MM--JDM integrates complementary modalities, skeleton sequences, optical flow, and structured textual descriptions, thereby providing richer and more discriminative representations of movement quality. This multi-modal design establishes a more comprehensive and challenging benchmark for evaluating AQA models and advancing research on multi-modal fusion and alignment.

\begin{figure}
    \centering
    \includegraphics[width=0.8\linewidth]{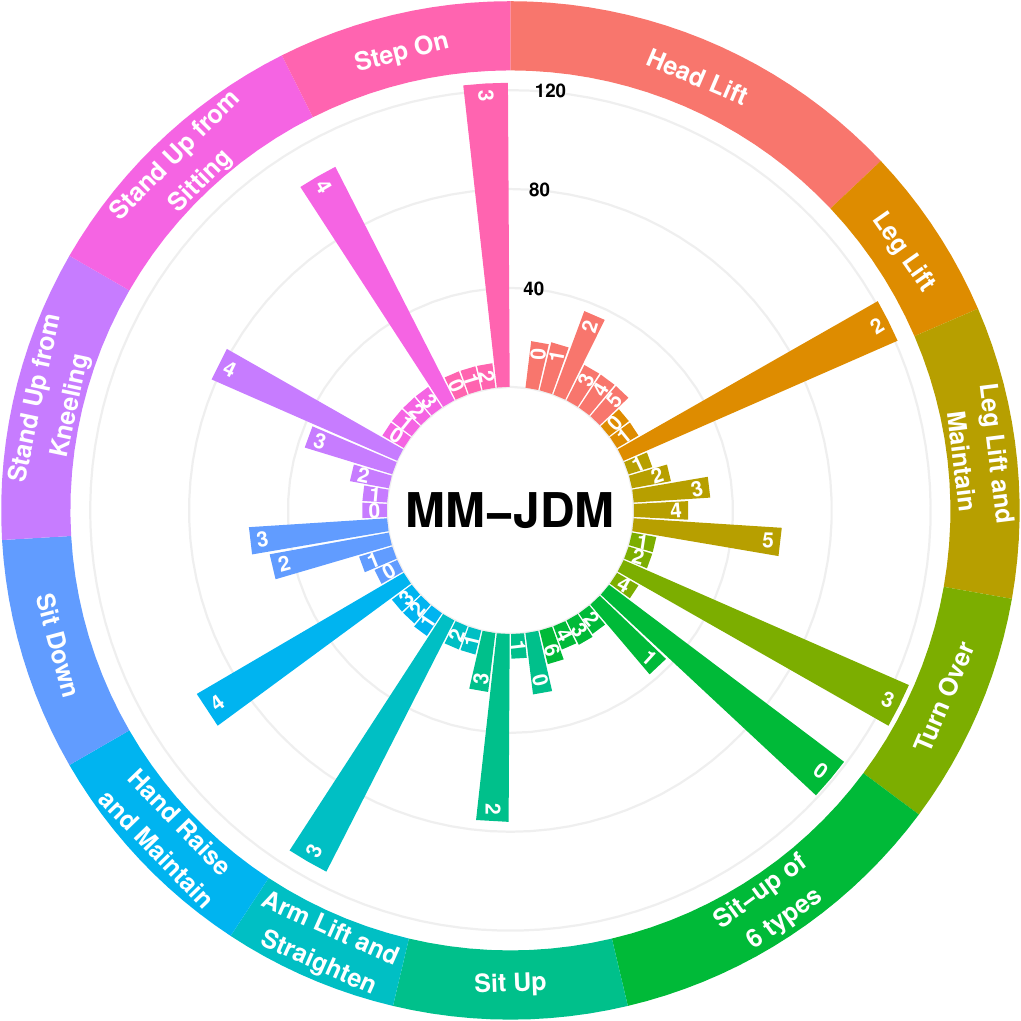}
    \caption{Category and sub-grade distribution of the MM--JDM dataset.}
    \label{fig:dataset}
\end{figure}

\subsection{Data Acquisition}
\label{sec:dataset_construction}
MM--JDM was constructed from recordings of children diagnosed with JDM at the Children's Hospital affiliated with the Capital Institute of Pediatrics. 
All data collection procedures followed institutional ethical guidelines, and informed consent was obtained from participants and their families. To protect privacy, the released videos were further anonymized through privacy-preserving preprocessing such as blurring. 

\myPara{Visual Data}
Recordings were captured using a variety of cameras and mobile devices during routine assessments, providing ecologically valid variations in illumination, viewpoint, and background. To ensure consistency across devices, all videos were standardized to a spatial resolution of $512 \times 512$. Three viewpoints (front, side, and top) were collected for each sample to provide comprehensive spatial coverage.
Skeleton sequences were extracted using the top-down HRNet-W32 model from MMPose~\cite{mmpose2020}, initialized with COCO pre-trained weights.  This provides structured joint-location information that enhances modeling of kinematic patterns. 
Optical flow was computed using the Farnebäck method~\cite{farneback2003two} to capture coarse motion dynamics complementary to RGB appearance. This lightweight approach provides stable and computationally efficient motion cues for temporal modeling. 

\myPara{Textual Data}
The textual modality includes structured descriptions of performed actions, physical examinations, and qualitative clinician comments. To ensure consistency in phrasing and reduce annotation burden, GPT-4o was used to generate initial draft descriptions based on non-privacy-sensitive metadata. These drafts followed a fixed template of the form: ``This is the action of [action type]. [Action description]. The subject of the video is [subject description]. [Physical examination notes].’’ The template constrains linguistic variability and reduces the risk of unintended semantic drift.
All generated text was reviewed and corrected by pediatric specialists to ensure accuracy and prevent medically misleading statements. The finalized descriptions do not contain severity-related cues that are not observable from the video, preventing label leakage and ensuring that textual information serves as a complementary rather than confounding modality. 
The skeleton and flow representations were verified as a sanity check to ensure accurate extraction. No manual correction, filtering, or relabeling was applied, and all representations were generated automatically. 

\subsection{Annotation Protocol}
A team of clinicians and researchers with expertise in JDM and pediatric motor assessment performed annotation. For each sample, action labels, movement-quality grades, and textual descriptions were independently reviewed. Disagreements were resolved through consensus discussions to ensure consistency. The skeleton and flow representations were additionally verified through qualitative checks to ensure accurate extraction of movement trajectories and temporal patterns, without any manual correction.

\subsection{Dataset Statistics} \label{sec:dataset_stat}
MM--JDM contains 1639 samples across 12 clinically relevant actions commonly performed during motor-function assessment. \cref{fig:dataset} illustrates the distribution of samples and sub-grades across actions.
The dataset includes recordings from 29 pediatric subjects, ranging from 3 to 15 years of age (mean $9.0 \pm 3.3$ years), with an approximately even gender distribution. Data were collected at multiple points in each subject's disease course, including early, mid-treatment, and follow-up visits, providing meaningful variability in motor performance.
The cohort exhibits a broad range of movement-quality grades across all actions, capturing differences in muscle strength, joint mobility, and execution quality.
Detailed grade distributions and sequence-length ranges are summarized in \cref{tab:MM-JDM-stat}. This diversity makes MM--JDM a challenging and representative benchmark for evaluating multi-modal AQA models and for studying alignment strategies under realistic clinical variability, as reflected by the substantial performance degradation of existing multi-modal methods in \cref{tab:exp}.
Although the dataset is designed primarily for research in automated movement-quality assessment, its multi-faceted annotations may ultimately contribute to the development of tools that support rehabilitation monitoring and clinical decision making.

\begin{table}[!t]
    \centering
    \caption{Detailed statistics of the MM--JDM dataset.}
    \setlength{\tabcolsep}{3pt}
    \resizebox{\linewidth}{!}{
        \begin{tabular}{clm{1.6cm}<\centering m{1.2cm}<\centering c}
            \toprule
            \textbf{Index}         & \textbf{Action Type}    & \textbf{Frames} & \textbf{Samples} & \textbf{Range} \\
            \midrule
            \rowcolor{orange!8} 01 & Head Lift               & 59--3960                 & 123                    & 0--5           \\
            \rowcolor{yellow!8} 02 & Leg Lift                & 33--\phantom{0}961       & 134                    & 0--2           \\
            \rowcolor{brown!8} 03  & Leg Lift and Maintain   & 90--3810                 & 136                    & 0--5           \\
            \rowcolor{orange!8} 04 & Turn Over               & 30--1169                 & 140                    & 0--4           \\
            \rowcolor{yellow!8} 05 & Sit Up of 6 types       & 30--1650                 & 193                    & 0--6           \\
            \rowcolor{brown!8} 06  & Sit Up                  & 30--1530                 & 133                    & 0--3           \\
            \rowcolor{orange!8} 07 & Arm Lift and Straighten & 30--\phantom{0}357       & 128                    & 0--3           \\
            \rowcolor{yellow!8} 08 & Hand Raise and Maintain & 59--2460                 & 126                    & 0--4           \\
            \rowcolor{brown!8} 09  & Sit Down                & 30--\phantom{0}570       & 127                    & 0--3           \\
            \rowcolor{orange!8} 10 & Stand Up from Kneeling  & 29--1407                 & 143                    & 0--4           \\
            \rowcolor{yellow!8} 11 & Stand Up from Sitting   & 30--1204                 & 123                    & 0--4           \\
            \rowcolor{brown!8} 12  & Step On                 & 29--\phantom{0}398       & 133                    & 0--3           \\
            \bottomrule
        \end{tabular}
    }
    \label{tab:MM-JDM-stat}
\end{table}

\begin{table*}
    \centering
    \caption{Comparison results with state-of-the-art methods on the MM--JDM dataset. 
    The best results are in \textbf{bold}, while the second-best results are \underline{underlined}.}
    \label{tab:exp}
    \resizebox{\linewidth}{!}{
    \begin{tabular}{crrcccccccccccccc}
        \toprule
        \multirow{2.5}{*}{\textbf{Type}} & 
        \multirow{2.5}{*}{\textbf{Methods}} & 
        \multirow{2.5}{*}{\textbf{Publisher}} &
        \multicolumn{13}{c}{\textbf{MM--JDM}} \\
        \cmidrule(lr){4-16}
         & & & 
         {01} & {02} & {03} & {04} & {05} & {06} & {07} & {08} & 
         {09} & {10} & {11} & {12} & \textbf{Avg.} \\
        \midrule

        \multicolumn{3}{c}{} & \multicolumn{13}{c}{\textbf{Spearman's Rank Correlation Coefficient ($\times$100) $\uparrow$}}  \\
        \cmidrule(lr){4-16}

\rowcolor{orange!5} Unimodal & CoRe \cite{yu2021group} & ICCV'21 
        & 68.14 & \underline{51.46} & 74.67 & 48.10 & 71.19 & 74.49 & 55.11 & 55.66 & 76.06 & \textbf{82.26} & 59.20 & 37.77 & 62.84 \\

        \rowcolor{orange!5} Unimodal & GDLT \cite{xu2022likert} & CVPR'22 
        & 68.07 & 47.40 & \underline{75.73} & 42.50 & 65.40 & 72.55 & 48.03 & \underline{58.57} & 72.32 & 66.13 & 53.66 & 36.80 & 58.93 \\

        \rowcolor{orange!5} Unimodal & HGCN \cite{zhou2023hierarchical} & TCSVT'23
        & 71.01 & \underline{51.46} & 65.39 & 50.71 & 70.95 & 79.75 & 51.80 & 55.18 & 79.99 & 76.15 & 63.47 & 37.77 & 62.80 \\

        \rowcolor{orange!5} Unimodal & DAE \cite{zhang2024auto} & NCAA'24
        & 74.19 & \underline{51.46} & 73.11 & 48.04 & \underline{72.69} & \underline{80.76} & 50.89 & 53.05 & 83.17 & \underline{82.16} & \underline{64.04} & 37.77 & \underline{64.28} \\

        \rowcolor{orange!5} Unimodal & T$^2$CR \cite{ke2024two} & INFS‘24
        & \underline{75.91} & \underline{51.46} & 64.16 & 50.71 & 72.57 & 75.16 & 48.22 & 50.30 & \underline{84.19} & 77.48 & 61.53 & 37.77 & 62.46 \\

        \rowcolor{orange!5} Unimodal & CoFInAl \cite{zhou2024cofinal} & IJCAI'24
        & 70.12 & 48.75 & 67.97 & 42.63 & 59.79 & 79.56 & 52.77 & 44.56 & 80.46 & 69.33 & 63.97 & 37.77 & 59.81 \\
        \rowcolor{orange!5} Unimodal & PHI \cite{zhou2025phi} & TIP'25 & 63.78 & 39.14 & 28.42 & 18.45 & 61.39 & 76.07 & 21.07 & 46.05 & 65.41 & 32.10 & 66.07 & 28.76 & 40.11 \\

\rowcolor{yellow!5} Multi-modal & RICA$^2$ \cite{majeedi2024rica} & ECCV'24
        & 45.45 & 40.37 & 53.68 & 39.27 & 34.21 & 31.96 & 41.98 & 42.43 & 45.96 & 39.40 & 45.90 & 33.70 & 45.81 \\

        \rowcolor{yellow!5} Multi-modal & MVLA \cite{xu2024vision} & ECCV'24
        & 62.86 & \underline{51.46} & 63.62 & \underline{50.91} & 67.56 & 66.81 & \underline{56.36} & 52.27 & 79.72 & 80.72 & 53.48 & 37.77 & 60.30 \\

        \rowcolor{yellow!5} Multi-modal & PAMFN \cite{zeng2024multimodal} & TIP'24
        & 64.64 & 49.98 & 56.76 & 49.80 & 68.72 & 60.16 & 37.73 & 32.59 & 30.84 & 67.62 & 39.95 & \underline{50.41} & 50.77 \\

        \rowcolor{yellow!5} Multi-modal & MLAVL \cite{xu2025language} & CVPR'25  & 62.25 & 65.19 & 45.51 & 33.74 & 60.52 & 68.37 & 30.68 & 31.33 & 58.05 & 77.77 & 43.93 & 43.78 & 51.76 \\

        \rowcolor{yellow!5} Multi-modal & Ours & --
        & \textbf{93.81} & \textbf{92.38} & \textbf{81.03} & \textbf{90.44} & \textbf{87.87} & \textbf{82.80} & \textbf{85.29} & \textbf{78.15} & \textbf{90.17} & 70.56 & \textbf{72.77} & \textbf{100.0} & \textbf{85.44} \\

\midrule
\multicolumn{3}{c}{} & \multicolumn{13}{c}{\textbf{relative Mean Square Error (rMSE) $\downarrow$}} \\
\cmidrule(lr){4-16}

\rowcolor{orange!5} Unimodal & CoRe \cite{yu2021group} & ICCV'21
& \phantom{0}7.216 & 1.293 & \phantom{0}5.772 & 5.729 & \underline{2.050} & 2.990 & 3.116 & \phantom{0}2.055 & \phantom{0}4.830 & \underline{2.209} & 1.301 & \underline{0.248} & 3.234 \\

\rowcolor{orange!5} Unimodal & GDLT \cite{xu2022likert} & CVPR'22
& \phantom{0}7.943 & 1.602 & \phantom{0}\underline{4.953} & 4.600
& \textbf{1.955} & 3.206 & 5.550 & \phantom{0}\underline{2.038}
& \phantom{0}4.904 & 3.320 & 2.055 & 1.223 & 3.612 \\

\rowcolor{orange!5} Unimodal & HGCN \cite{zhou2023hierarchical} & TCSVT'23
& \phantom{0}6.911 & 0.820 & \phantom{0}7.088 & 4.512
& 2.305 & 2.378 & \underline{2.881} & \phantom{0}2.390
& \phantom{0}3.906 & 3.041 & 1.809 & 1.202 & 3.270 \\

\rowcolor{orange!5} Unimodal & DAE \cite{zhang2024auto} & NCAA'24
& \phantom{0}6.100 & \underline{0.804} & \phantom{0}6.066 & 4.267
& 2.693 & \textbf{1.989} & 3.328 & \phantom{0}3.277
& \phantom{0}3.949 & \textbf{2.098} & 1.425 & 0.433 & \underline{3.036} \\

\rowcolor{orange!5} Unimodal & T$^2$CR \cite{ke2024two} & INFS‘24
& \phantom{0}\underline{5.567} & \textbf{0.533} & \phantom{0}7.727 & 4.102
& 2.239 & 3.378 & 4.080 & \phantom{0}2.858
& \phantom{0}\underline{3.231} & 2.442 & 1.628 & 0.635 & 3.202 \\

\rowcolor{orange!5} Unimodal & CoFInAl \cite{zhou2024cofinal} & IJCAI'24
& \phantom{0}6.801 & 1.178 & \phantom{0}6.321 & 3.301
& 2.905 & \underline{2.078} & 3.271 & \phantom{0}3.145
& \phantom{0}4.651 & 3.386 & \underline{1.018} & 0.492 & 3.212 \\

\rowcolor{orange!5} Unimodal & PHI \cite{zhou2025phi} & TIP'25 & \phantom{0}9.816 & 4.103 & 20.357 & 7.934 & 3.652 & 3.806 & 8.531 & 15.445 & \phantom{0}7.781 & 8.312 & 5.654 & 4.018 & 8.284 \\

\rowcolor{yellow!5} Multi-modal & RICA$^2$ \cite{majeedi2024rica} & ECCV'24
& 11.523 & 3.655 & 10.872 & 6.722 & 6.814 & 7.728
& 7.496 & \phantom{0}5.183 & \phantom{0}9.460 & 4.815 & 5.920 & 3.172 & 6.947 \\

\rowcolor{yellow!5} Multi-modal & MVLA \cite{xu2024vision} & ECCV'24
& \phantom{0}8.286 & 1.098 & \phantom{0}8.189 & 4.017
& 2.919 & 4.212 & 5.670 & \phantom{0}2.956
& \phantom{0}4.647 & 2.763 & 3.883 & 1.538 & 4.182 \\

\rowcolor{yellow!5} Multi-modal & PAMFN \cite{zeng2024multimodal} & TIP'24
& \phantom{0}8.779 & 6.597 & 12.241 & \textbf{1.547}
& 4.742 & 6.124 & 3.766 & \phantom{0}9.437
& 10.936 & 4.331 & 6.565 & 3.026 & 6.508 \\

\rowcolor{yellow!5} Multi-modal & MLAVL \cite{xu2025language} & CVPR'25  & \phantom{0}6.753 & 2.724 & 11.655 & 2.559 & 5.949 & 6.290 & 6.087 & 12.668 & \phantom{0}7.940 & 3.208 & 5.083 & 2.972 & 6.157 \\

\rowcolor{yellow!5} Multi-modal & Ours & --
& \phantom{0}\textbf{1.333} & 1.250 & \phantom{0}\textbf{3.594} & \underline{2.976}
& 3.353 & 5.128 & \textbf{1.316} & \phantom{0}\textbf{1.201}
& \phantom{0}\textbf{3.216} & 4.464 & \textbf{0.521} & \textbf{0.000} & \textbf{2.363} \\

        \midrule
        \multicolumn{3}{c}{} & \multicolumn{13}{c}{\textbf{Accuracy (\%) $\uparrow$}} \\   
        \cmidrule(lr){4-16}

\rowcolor{orange!5} Unimodal & CoRe \cite{yu2021group} & ICCV'21
        & 35.14 & \underline{95.12} & 56.10 & \underline{92.86} & 79.31 & 85.00 & \underline{94.87} & 89.47 & 69.23 & \textbf{83.72} & \underline{89.19} & \underline{97.50} & 80.63 \\

        \rowcolor{orange!5} Unimodal & GDLT \cite{xu2022likert} & CVPR'22
        & 40.54 & 92.68 & \underline{60.98} & \underline{92.86} & 79.31 & 75.00 & 87.18 & \underline{92.11} & 65.40 & 67.44 & 83.78 & 95.00 & 77.69 \\

        \rowcolor{orange!5} Unimodal & HGCN \cite{zhou2023hierarchical} & TCSVT'23
        & 40.54 & \textbf{97.56} & 58.54 & \underline{92.86} & \underline{81.03} & \underline{87.50} & \underline{94.87} & 89.47 & \underline{74.36} & \underline{81.40} & 86.49 & 95.00 & \underline{81.64} \\

        \rowcolor{orange!5} Unimodal & DAE \cite{zhang2024auto} & NCAA'24
        & 40.54 & \textbf{97.56} & 46.34 & \underline{92.86} & 77.59 & \textbf{90.00} & 89.74 & 78.95 & 69.23 & 79.07 & 83.78 & \underline{97.50} & 78.60 \\

        \rowcolor{orange!5} Unimodal & T$^2$CR \cite{ke2024two} & INFS‘24
        & 37.84 & \textbf{97.56} & 53.66 & \underline{92.86} & 75.86 & 82.50 & \underline{94.87} & 84.21 & \underline{74.36} & 79.07 & 86.49 & \underline{97.50} & 79.73 \\

        \rowcolor{orange!5} Unimodal & CoFInAl \cite{zhou2024cofinal} & IJCAI'24
        & \underline{48.65} & 92.68 & 51.22 & \textbf{95.24} & 17.24 & 85.00 & 92.31 & 89.47 & \underline{74.36} & 79.07 & \underline{89.19} & \underline{97.50} & 75.99 \\

        \rowcolor{orange!5} Unimodal & PHI \cite{zhou2025phi} & TIP'25 & 35.14 & 90.24 & 40.00 & 90.48 & 74.14 & 82.50 & 87.18 & 59.45 & 43.59 & 30.23 & 54.76 & 95.00 & 65.23 \\

\rowcolor{yellow!5} Multi-modal & RICA$^2$ \cite{majeedi2024rica} & ECCV'24
        & 24.32 & 90.24 & 48.78 & 90.48 & 72.41 & 67.50 & 87.18 & 86.84 & 48.72 & 60.47 & 83.78 & 95.00 & 71.31 \\

        \rowcolor{yellow!5} Multi-modal & MVLA \cite{xu2024vision} & ECCV'24
        & 37.84 & \textbf{97.56} & 56.10 & \underline{92.86} & 74.14 & 70.00 & 89.74 & 84.21 & 66.67 & \textbf{83.72} & 83.78 & 95.00 & 77.64 \\

        \rowcolor{yellow!5} Multi-modal & PAMFN \cite{zeng2024multimodal} & TIP'24
        & 33.33 & 85.00 & 45.00 & \underline{92.86} & 71.93 & 58.97 & 92.11 & 70.27 & 52.63 & 66.67 & 88.89 & 92.31 & 70.83 \\

        \rowcolor{yellow!5} Multi-modal & MLAVL \cite{xu2025language} & CVPR'25 & 36.11 & 90.00 & 17.50 & 76.19 & 63.16 & 48.72 & 68.42 & 35.14 & 26.32 & 54.76 & 77.78 & 41.03 & 52.93 \\

        \rowcolor{yellow!5} Multi-modal & Ours & --
        &  \textbf{91.67} & 95.00 & \textbf{72.50} & 88.10 & \textbf{87.93} & 64.10 & \textbf{97.37} & \textbf{94.74} & \textbf{86.84} & 76.19 & \textbf{94.44} & \textbf{100.0} & \textbf{87.41} \\

        \bottomrule
    \end{tabular}
    }
\end{table*}

\section{Experiments} \label{sec:experiments}
In this section, we first describe the external datasets, evaluation metrics, and implementation details in \cref{sec:setting}. Next, \cref{sec:comparison} compares our method with state-of-the-art baselines to demonstrate its effectiveness and generalization across diverse multi-modal AQA benchmarks. We conduct ablation studies in \cref{sec:ablation} to analyze the contributions of the different components in DualAlign. Finally, \cref{sec:qq} provides a series of quantitative and qualitative analyses of the model's performance.

\subsection{Experimental Setting} \label{sec:setting}
\myPara{Additional Datasets for Generalization Evaluation}
To further examine the generalization ability of our method beyond the proposed MM--JDM dataset, we additionally evaluate it on two widely used AQA datasets. Although these datasets were originally unimodal \cite{zeng2020hybrid,parmar2017learning}, some recent studies \cite{xu2024vision,xu2025language,du2023learning} have explored incorporating auxiliary modalities for analysis. To maintain consistency across all experiments, we construct multi-modal counterparts for RG and FIS-V by applying the same processing procedures as in the MM--JDM setting. 
Specifically, the fields [Action description] and [subject description] (see \cref{sec:dataset_construction}) are constructed using a template-based strategy, based on available metadata and prior descriptions from \cite{du2023learning}, while the physical examination field is left empty.
These descriptions (see \cref{fig:case_study-rg}) provide general semantic context and do not include any grade- or label-related information, thereby avoiding potential label leakage. The \textbf{Rhythmic Gymnastics (RG)} dataset \cite{zeng2020hybrid} contains 1,000 videos covering four rhythmic gymnastics apparatuses (Ball, Clubs, Hoop, and Ribbon), with each video lasting approximately 1.6 minutes at 25 FPS. According to the official protocol, 200 videos per category are used for training and 50 for evaluation. The \textbf{Figure Skating Video (FIS-V)} dataset \cite{xu2019learning} includes 500 videos of ladies’ singles short programs, each about 2.9 minutes long and recorded at 25 FPS. Following the standard split, 400 videos are used for training and 100 for testing, and each video contains two official scores, namely the Total Element Score (TES) and the Total Program Component Score (PCS). We train separate models for each score following \citet{xu2019learning}.

\myPara{Evaluation Metrics}
We evaluate the performance of all models using three standard metrics: Spearman's Rank Correlation Coefficient (SRCC), relative Mean Squared Error (rMSE), and accuracy. SRCC measures the correlation between predicted and ground-truth grades, providing insights into the model's ranking performance, which is:
\begin{equation} \label{eq:srcc}
    \text{SRCC} = \frac{ \sum_{i=1}^{N} \left( r_i - \bar{r} \right) \left( \hat{r}_i - \bar{\hat{r}} \right) }{ \sqrt{\sum_{i=1}^{N} \left( r_i - \bar{r} \right)^2 } \sqrt{\sum_{i=1}^{N} \left( \hat{r}_i - \bar{\hat{r}} \right)^2 } },
\end{equation}
where $r_i$ and $\hat{r}_i$ are ranks of the ground-truth and predicted grades, respectively, $\bar{r}$ and $\bar{\hat{r}}$ are the mean values of the ground-truth and predicted grade ranks, respectively, and $N$ is the sample size.
rMSE quantifies the prediction error in terms of grade values, while accuracy assesses the model's ability to predict the correct grade category, which is:
\begin{equation} \label{eq:rmse}
    \text{rMSE} = \frac{1}{N} \sum_{i=1}^{N} \left( \frac{\hat{y}_i - y_i}{y_{\max} - y_{\min}} \right)^2 \times 100,
\end{equation}
where $y_{\max}$ and $y_{\min}$ denote the maximum and minimum grades, respectively. The multiplication by 100 in \cref{eq:rmse} converts rMSE to a proper percentage scale. Accuracy is calculated as the percentage of correctly predicted grade categories, providing insights into the model's grading performance, which is:
\begin{equation} \label{eq:acc}
    \text{Accuracy} = \frac{1}{N} \sum_{i=1}^{N} \mathbbm{1} \left(  \hat{y}_i = y_i  \right) \times 100,
\end{equation}
where $\mathbbm{1}(\cdot)$ is the indicator function.
These metrics offer a holistic evaluation of the model's performance.

\myPara{Implementation Details}
The feature dimensions $D_1$ and $D_2$ are set to 1024 and 512, respectively. We employ a VST backbone \cite{liu2022video} and an I3D backbone \cite{carreira2017quo} pre-trained on the Kinetics-400 dataset \cite{kay2017kinetics} as the encoders to extract features from video and optical flow data, respectively.
To address sequence–length variability, we uniformly sample each video to 103 frames, which ensures full temporal coverage of the action while reducing variation across samples. This design is consistent with prior AQA protocols~\cite{yu2021group,zhou2023hierarchical}, where fixed-length sampling has been shown to stabilize training under variable sequence lengths. In contrast, more flexible strategies such as sliding windows may introduce additional variability in small-scale settings and are less commonly adopted in standard AQA benchmarks.  Moreover, consistent with prevailing AQA practice \cite{xu2022likert,zeng2020hybrid}, we adopt an action-specific modeling setting to mitigate the confounding effects caused by cross-action discrepancies in scoring criteria, score ranges, and motion characteristics.
Our model is implemented using PyTorch and trained on a single RTX 3090 GPU. 
We use the stochastic gradient descent (SGD) optimizer for training. The learning rate is initialized to $1 \times 10^{-2}$ and is gradually decayed using a cosine annealing schedule to facilitate stable and smooth convergence.
To further regularize the model and prevent overfitting, we apply a dropout rate of 0.5, temperature $\tau$ of 0.1, and weight decay of $1 \times 10^{-4}$. The loss weights $\lambda_1$ and $\lambda_2$ are both set to 1, as the two losses are normalized and defined on consistent scales, allowing them to contribute equally to the optimization. Additionally, cross-attention is employed within the enhanced modules to capture inter-modal dependencies and improve feature alignment.
Unless otherwise specified, baseline methods are reproduced following their official implementations and original hyper-parameter settings, including backbone choice, optimizer type, learning rate schedule, and training epochs. When direct transfer to MM--JDM is not feasible (e.g., due to input dimensions or action-specific prediction heads), we adapt the implementation while preserving the original model design. All baselines use the same dataset splits, input preprocessing, and evaluation protocol to ensure fair comparison.

\subsection{Comparison with the State-of-the-Art}
\label{sec:comparison}
We compare our DualAlign method with several state-of-the-art baselines, including both unimodal and multi-modal approaches. The unimodal baselines include CoRe \cite{yu2021group}, GDLT \cite{xu2022likert}, HGCN \cite{zhou2023hierarchical}, DAE \cite{zhang2024auto}, T$^2$CR \cite{ke2024two}, CoFInAl \cite{zhou2024cofinal}, and PHI \cite{zhou2025phi}. The multi-modal baselines include RICA$^2$ \cite{majeedi2024rica}, MVLA \cite{xu2024vision}, PAMFN \cite{zeng2024multimodal}, and MLAVL \cite{xu2025language}. These baselines represent the state-of-the-art in AQA and provide a comprehensive benchmark for evaluating our DualAlign method. 
All methods are evaluated under their original modality settings as described in their respective works, without restricting their input modalities (e.g., audio in MLAVL \cite{xu2025language}). This avoids artificially limiting any method and reflects a realistic comparison where each approach operates as designed. 

\begin{table}[]
    \centering
    \setlength{\tabcolsep}{3pt}
    \caption{Comparison results on the RG dataset. The best and second-best results are highlighted in \textbf{bold} and \underline{underlined}, respectively.}
    \label{tab:rg}
    \resizebox{\linewidth}{!}{
    \begin{tabular}{crrccccc}
    \toprule
    \multirow{2.5}{*}{\textbf{Type}} & 
    \multirow{2.5}{*}{\textbf{Methods}} & 
    \multirow{2.5}{*}{\textbf{Publisher}} &
    \multicolumn{5}{c}{\textbf{SRCC} $\uparrow$}  \\ \cmidrule(lr){4-8}
    & & & Ball & Clubs & Hoop & Ribbon & Avg. \\
    \midrule
    \rowcolor{orange!8}Unimodal & CoRe \cite{yu2021group} & ICCV’21 & 0.613 & 0.749 & 0.749 & 0.688 & 0.704 \\
    \rowcolor{orange!8}Unimodal & GDLT \cite{xu2022likert} & CVPR'22 & 0.746 & 0.802 & 0.765 & 0.741 & 0.765 \\
    \rowcolor{orange!8}Unimodal & CoFInAl \cite{zhou2024cofinal} & IJCAI'24 & 0.809 & 0.806 & 0.804 & 0.810 & 0.807 \\
    \rowcolor{orange!8}Unimodal & PHI \cite{zhou2025phi} & TIP'25 & 0.818 & 0.803 & 0.812 & 0.805 & 0.810 \\
    \rowcolor{yellow!8}Multi-modal & PAMFN \cite{zeng2024multimodal} & TIP'24 & 0.757 & 0.825 & 0.836 & 0.846 & 0.819  \\
    \rowcolor{yellow!8}Multi-modal & MLAVL \cite{xu2025language} & CVPR'25 & \underline{0.826} & \underline{0.829} & \underline{0.871} & \underline{0.866} & \underline{0.849}  \\
    \rowcolor{yellow!8}Multi-modal & Ours & -- & \bf 0.857 & \bf 0.871 & \textbf{0.875} & \bf 0.908 & \bf 0.878 \\
    \midrule
    & & & 
    \multicolumn{5}{c}{\textbf{MSE} $\downarrow$}  \\ \cmidrule(lr){4-8}
    \rowcolor{orange!8}Unimodal & CoRe \cite{yu2021group} & ICCV’21 & 9.19 & 4.83 & 5.26 & 7.43 & 6.68 \\
    \rowcolor{orange!8}Unimodal & GDLT \cite{xu2022likert} & CVPR'22 & 5.90 & 4.34 & 5.70 & 6.16 & 5.53 \\
    \rowcolor{orange!8}Unimodal & CoFInAl \cite{zhou2024cofinal} & IJCAI'24 & \underline{5.07} & 5.19 & 6.37 & 6.30 & 5.73 \\
    \rowcolor{orange!8}Unimodal & PHI \cite{zhou2025phi} & TIP'25 & 8.21 & 4.68 & 5.63 & 7.40 & 6.48 \\
    \rowcolor{yellow!8}Multi-modal & PAMFN \cite{zeng2024multimodal} & TIP'24 & 6.24 & 7.45 & 5.21 & 7.67 & 6.64  \\
    \rowcolor{yellow!8}Multi-modal & MLAVL \cite{xu2025language} & CVPR'25 & 5.57 & \underline{4.20} & \bf 4.11 & \underline{3.99} & \underline{4.47}  \\
    \rowcolor{yellow!8}Multi-modal & Ours  & -- & \textbf{4.65} & \bf 3.67 & \underline{4.36} & \bf 2.61 & \bf 3.82 \\
    \midrule
    & & & 
    \multicolumn{5}{c}{\textbf{rMSE} $\downarrow$}  \\ \cmidrule(lr){4-8}
    \rowcolor{orange!8}Unimodal & CoRe \cite{yu2021group} & ICCV’21 
    & 2.45 & 2.26 & 1.98 & 2.78 & 2.37 \\
    
    \rowcolor{orange!8}Unimodal & GDLT \cite{xu2022likert} & CVPR'22 
    & 2.39 & 3.12 & 2.38 & 2.05 & 2.49 \\
    
    \rowcolor{orange!8}Unimodal & CoFInAl \cite{zhou2024cofinal} & IJCAI'24 
    & \underline{1.36} & 2.45 & 9.92 & 2.38 & 4.03 \\
    
    \rowcolor{orange!8}Unimodal & PHI \cite{zhou2025phi} & TIP'25 
    & 2.19 & 2.19 & 2.12 & 2.77 & 3.00 \\
    
    \rowcolor{yellow!8}Multi-modal & PAMFN \cite{zeng2024multimodal} & TIP'24 
    & 1.66 & 3.49 & 1.96 & 2.87 & 2.50 \\
    
    \rowcolor{yellow!8}Multi-modal & MLAVL \cite{xu2025language} & CVPR'25 
    & 1.48 & \underline{1.97} & \bf 1.55 & \underline{1.49} & \underline{1.62} \\
    
    \rowcolor{yellow!8}Multi-modal & Ours  & -- & \textbf{1.24} & \bf 1.72 & \underline{1.64} & \bf 1.04 & \bf 1.41 \\
    \bottomrule
    \end{tabular}
    }
\end{table}

\myPara{Assessment Performance on MM--JDM}
The overall assessment results are summarized in \cref{tab:exp}. On the MM--JDM dataset, DualAlign achieves the highest average SRCC ($\times 100$) of 85.44, outperforming the best unimodal baseline DAE~\cite{zhang2024auto} (64.28) and the strongest multi-modal baseline MVLA~\cite{xu2024vision} (60.30) by more than 21 and 25 points, respectively. In terms of error, our method attains the lowest average rMSE of 2.363, which corresponds to a reduction of about 22\% relative to DAE (3.036) and an even larger margin over recent multi-modal competitors such as RICA$^2$~\cite{majeedi2024rica} (6.947). Accuracy exhibits a consistent trend: DualAlign reaches an average of 87.41\%, substantially higher than the 80\% level of the strongest baselines, indicating that the proposed alignment strategy provides a favorable trade-off between ranking consistency, regression precision, and discrete grading reliability. On many individual actions, the improvements are also pronounced. For example, on Actions~01 and 02, our SRCCs of 93.81 and 92.38 substantially exceed those of T$^2$CR~\cite{ke2024two} (75.91 and 51.46), and on Actions~03 and 07 our rMSE values of 3.594 and 1.316 are the lowest among all methods, highlighting the benefit of multi-modal alignment for complex whole-body movements. The perfect SRCC on Action 12 results from its highly ordered grading protocol and limited ambiguity, rather than information leakage, as textual descriptions exclude grade-related cues.
Meanwhile, the per-action results also reveal limitations and trade-offs. For instance, on Action~10, several unimodal methods, such as CoRe~\cite{yu2021group} and DAE~\cite{zhang2024auto}, achieve higher SRCC and Accuracy than DualAlign, although our rMSE remains within a comparable range. This action has a relatively narrow score range and less pronounced motion variation, which favors specialized unimodal regressors and reduces the advantage of multi-modal fusion. On Actions~04, 05, and 06, certain baselines obtain slightly lower rMSE while still exhibiting inferior SRCC, suggesting that they prioritize local error minimization at the expense of global ranking consistency. In contrast, DualAlign maintains a more balanced performance across all three metrics, which is desirable for AQA where both ordinal consistency and absolute error matter. When viewed alongside the imbalance analysis in \cref{fig:diversity}, these patterns indicate that DualAlign is particularly robust to actions with diverse grade distributions and complex motion, while its gains are more modest for simpler or weakly varying actions. Overall, the results confirm that our method significantly advances the state-of-the-art on MM--JDM, while also clarifying the scenarios in which its advantage is less pronounced, thereby providing a more reliable performance assessment.

\begin{table}[]
    \centering
    \setlength{\tabcolsep}{6pt}
    \caption{Comparison results on the Fis-V dataset. The best and second-best results are highlighted in \textbf{bold} and \underline{underlined}, respectively.}
    \label{tab:fisv}
    \resizebox{0.9\linewidth}{!}{
    \begin{tabular}{crrccc}
    \toprule
    \multirow{2.5}{*}{\textbf{Type}} & 
    \multirow{2.5}{*}{\textbf{Methods}} & 
    \multirow{2.5}{*}{\textbf{Publisher}} &
    \multicolumn{3}{c}{\textbf{SRCC} $\uparrow$}  \\ \cmidrule(lr){4-6}
    & & & TES & PCS & Avg. \\
    \midrule
    \rowcolor{orange!8}Unimodal & CoRe \cite{yu2021group} & ICCV’21 & 0.640 & 0.763 & 0.707 \\
    \rowcolor{orange!8}Unimodal & GDLT \cite{xu2022likert} & CVPR'22 & 0.667 & 0.822 & 0.755 \\
    \rowcolor{orange!8}Unimodal & CoFInAl \cite{zhou2024cofinal} & IJCAI'24 & 0.711 & 0.860 & 0.797 \\
    \rowcolor{orange!8}Unimodal & PHI \cite{zhou2025phi} & TIP'25 & 0.726 & 0.867 & 0.804 \\
    \rowcolor{yellow!8}Multi-modal & PAMFN \cite{zeng2024multimodal} & TIP'24 & 0.754 & \underline{0.872} & 0.822  \\
    \rowcolor{yellow!8}Multi-modal & MLAVL \cite{xu2025language} & CVPR'25 & \underline{0.766} & 0.863 & \underline{0.823}  \\
    \rowcolor{yellow!8}Multi-modal & Ours & -- & \textbf{0.844} & \textbf{0.895} & \textbf{0.872}  \\
    \midrule
    & & & 
    \multicolumn{3}{c}{\textbf{MSE} $\downarrow$}  \\ \cmidrule(lr){4-6}
    \rowcolor{orange!8}Unimodal & CoRe \cite{yu2021group} & ICCV’21 & 22.62 & 10.31 & 16.46 \\
    \rowcolor{orange!8}Unimodal & GDLT \cite{xu2022likert} & CVPR'22 & 30.06 & \phantom{0}8.93 & 19.50 \\
    \rowcolor{orange!8}Unimodal & CoFInAl \cite{zhou2024cofinal} & IJCAI'24 & 30.05 & 13.65 & 21.85 \\
    \rowcolor{orange!8}Unimodal & PHI \cite{zhou2025phi} & TIP'25 & 30.28 & 10.38 & 20.33 \\
    \rowcolor{yellow!8}Multi-modal & PAMFN \cite{zeng2024multimodal} & TIP'24 & 22.50 & \phantom{0}8.16 & 15.33  \\
    \rowcolor{yellow!8}Multi-modal & MLAVL \cite{xu2025language} & CVPR'25 & \underline{19.44} & \phantom{0}\underline{7.17} & \underline{13.31}  \\
    \rowcolor{yellow!8}Multi-modal & Ours & -- & \textbf{16.33} & \phantom{0}\textbf{6.98} & \textbf{11.66}  \\
    \midrule
    & & & 
    \multicolumn{3}{c}{\textbf{rMSE} $\downarrow$}  \\ \cmidrule(lr){4-6}
    \rowcolor{orange!8}Unimodal & CoRe \cite{yu2021group} & ICCV’21 & 1.90 & 1.65 & 1.77 \\
    \rowcolor{orange!8}Unimodal & GDLT \cite{xu2022likert} & CVPR'22 & 2.52 & 1.43 & 1.97 \\
    \rowcolor{orange!8}Unimodal & CoFInAl \cite{zhou2024cofinal} & IJCAI'24 & 2.52 & 2.18 & 2.35 \\
    \rowcolor{orange!8}Unimodal & PHI \cite{zhou2025phi} & TIP'25 & 2.54 & 1.66 & 2.18 \\
    \rowcolor{yellow!8}Multi-modal & PAMFN \cite{zeng2024multimodal} & TIP'24 & 1.89 & 1.31 & 1.60 \\
    \rowcolor{yellow!8}Multi-modal & MLAVL \cite{xu2025language} & CVPR'25 & \underline{1.63} & \underline{1.15} & \underline{1.39} \\
    \rowcolor{yellow!8}Multi-modal & Ours & -- & \textbf{1.37} & \textbf{1.11} & \textbf{1.24}  \\
    \bottomrule
    \end{tabular}
    }
\end{table}

\myPara{Comparison with Large Language Models on MM--JDM}
To provide a contemporary reference for fine-grained AQA, we additionally evaluate two strong multi-modal large language models (mLLMs), GPT-4o and Gemini-3 Pro, under a zero-shot setting on MM--JDM. For fair comparison, each model is prompted with the action category, the corresponding scoring rules (where a higher grade indicates easier and better action performance), and the action description text, and is required to output only one integer grade within the valid range of that action (see \cref{tab:MM-JDM-stat}). No few-shot examples or additional prompt engineering are used.
As shown in \cref{fig:gpt-waterfall}, both GPT-4o and Gemini-3 Pro exhibit highly unstable and inconsistent performance across different actions. Their average SRCC values are only 0.09 and -0.01, respectively, with significantly lower accuracy (20.29\% and 36.87\%) compared to DualAlign (87.41\%). This indicates that although mLLMs are effective at high-level semantic understanding and description generation, they struggle to capture subtle fine-grained differences required for quantitative action quality assessment, especially in medical scenarios where small posture variations correspond to different clinical grades.
In contrast, DualAlign achieves consistently strong performance by explicitly modeling cross-modal representation discrepancies under supervised learning. These results suggest that current general-purpose mLLMs are not yet reliable substitutes for specialized AQA models in fine-grained assessment tasks.

\begin{figure}
    \centering
    \includegraphics[width=\linewidth]{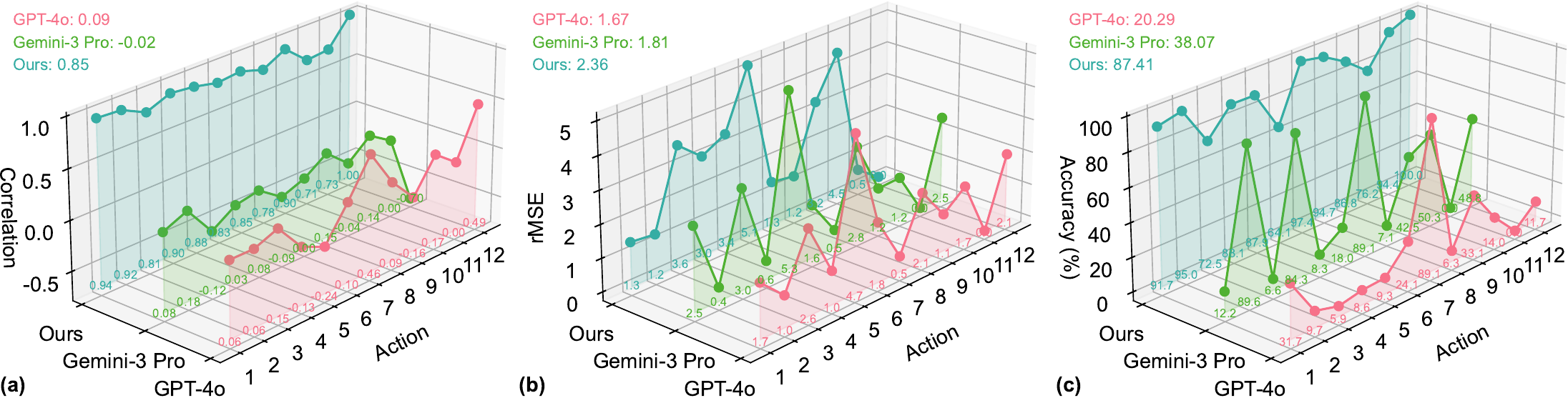}
    \MSfigcaption{fig:gpt-waterfall}{Zero-shot evaluation of GPT-4o, Gemini-3 Pro, and DualAlign on MM--JDM across 12 actions. (a) SRCC, (b) rMSE, and (c) Accuracy. While modern mLLMs show strong general visual understanding ability, they produce unstable and inconsistent grading results for fine-grained AQA.}{}
\end{figure} 

\myPara{External Validation and Cross-Domain Generalization}
To assess whether DualAlign generalizes beyond the self-constructed MM--JDM dataset, we further evaluate it on two widely used external AQA benchmarks, RG and FIS-V, which differ substantially from the medical domain in motion patterns, appearance statistics, and scoring granularity.  
Following the standard supervised AQA protocol, DualAlign is trained on the official training split of each dataset and evaluated on its corresponding test split, rather than being tested in a zero-shot cross-dataset setting. These experiments therefore examine the adaptability of DualAlign under standard in-domain benchmark protocols while validating whether the proposed discrepancy-aware alignment strategy remains effective across diverse AQA scenarios. As shown in \cref{tab:rg,tab:fisv}, DualAlign consistently achieves state-of-the-art performance. On RG, our method reaches an average SRCC of 0.878, outperforming the previous best approach by 3.4\% and achieving top performance in three of the four events, including an SRCC of 0.908 on Ribbon. It also reduces the average MSE by 14.5\% and rMSE by 13.0\%. On Fis-V, the improvements are even more pronounced: DualAlign attains an average SRCC of 0.872, surpassing the prior state of the art by 6.0\%, and boosts TES SRCC from 0.766 to 0.844 (a 10.2\% gain). The average MSE decreases from 13.31 to 11.66, reflecting a 12.4\% reduction. These results demonstrate that the proposed two-stage alignment produces stable and transferable cross-modal representations, enabling strong generalization to domains with fundamentally different motion statistics and scoring protocols.

\begin{figure}[!t]
    \centering
    \sf
    \includegraphics[width=\linewidth,clip,trim=60 250 50 220]{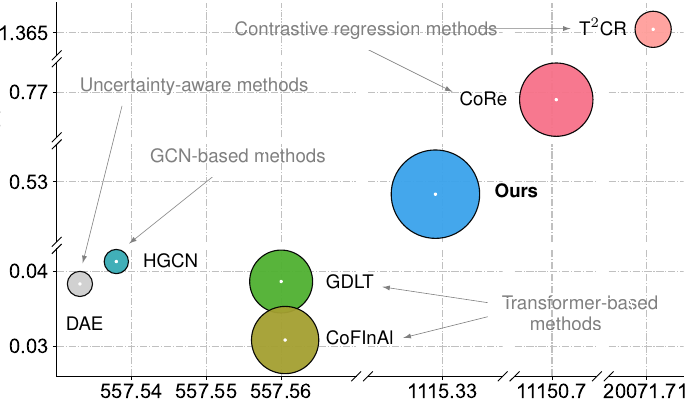}
    \caption{Computational performance comparison with selected baselines under the identical device settings on the MM--JDM dataset. The $x$-axis represents the FLOPs, the $y$-axis indicates the inference time, and the bubble size corresponds to the number of parameters.}
    \label{fig:bubble-plot}
\end{figure}

\myPara{Computational Performance}
To evaluate the computational efficiency of our DualAlign method, we compare it with selected baselines in terms of the number of parameters, FLOPs, and inference time. The results are presented in \cref{fig:bubble-plot}, where each bubble represents a method, with the $x$-axis indicating FLOPs, the $y$-axis representing inference time per video sample (103 frames),  and the bubble size corresponding to the number of parameters. 
Inference time is measured on the model forward pass only,  excluding preprocessing steps (e.g., optical flow and pose estimation), to isolate the computational cost of the model itself.
Our DualAlign method achieves a balance between computational complexity and performance, demonstrating superior efficiency compared to other methods.
For instance, it outperforms CoRe~\cite{yu2021group} and T$^2$CR~\cite{ke2024two} in terms of FLOPs (1115.3 GFLOPs v.s. 11150.7 GFLOPs and 20071.7 GFLOPs, respectively) and inference time (528 ms v.s. 769 ms and 1365 ms, respectively). Additionally, it consumes a smaller number of extra parameters compared to other methods, such as HGCN~\cite{zhou2023hierarchical} (28.35 M v.s. 12.56 M) and CoFInAl~\cite{zhou2024cofinal} (28.35 M v.s. 14.39 M).
This competitive computational efficiency ensures that our method is not only effective but also scalable for real-world applications.

\subsection{Ablation Study}
\label{sec:ablation}
\begin{figure}[]
    \centering
    \includegraphics[width=\linewidth]{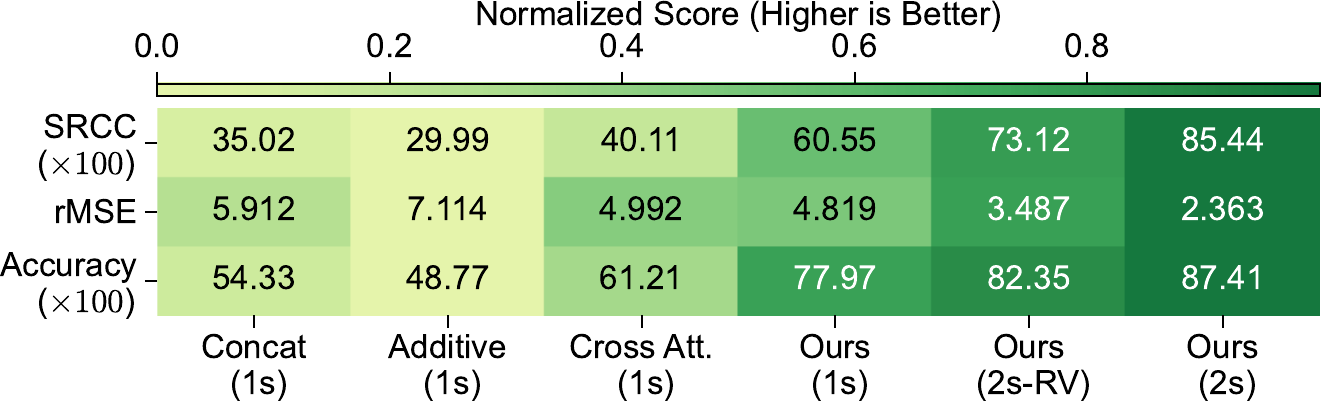}
    \caption{Normalized performance comparison of one-stage and two-stage fusion strategies.
    The heatmap visualizes SRCC, rMSE, and Accuracy scores scaled to the interval $[0,1]$ for direct comparison across methods.
    “1s" denotes one-stage fusion, “2s" denotes our two-stage DualAlign strategy, and “RV" denotes the reverse fusion order, where textual features are fused into RGB prior to the alignment with flow and skeleton features.}
    \label{fig:stage}
\end{figure}

\begin{figure}
    \centering
    \includegraphics[width=\linewidth]{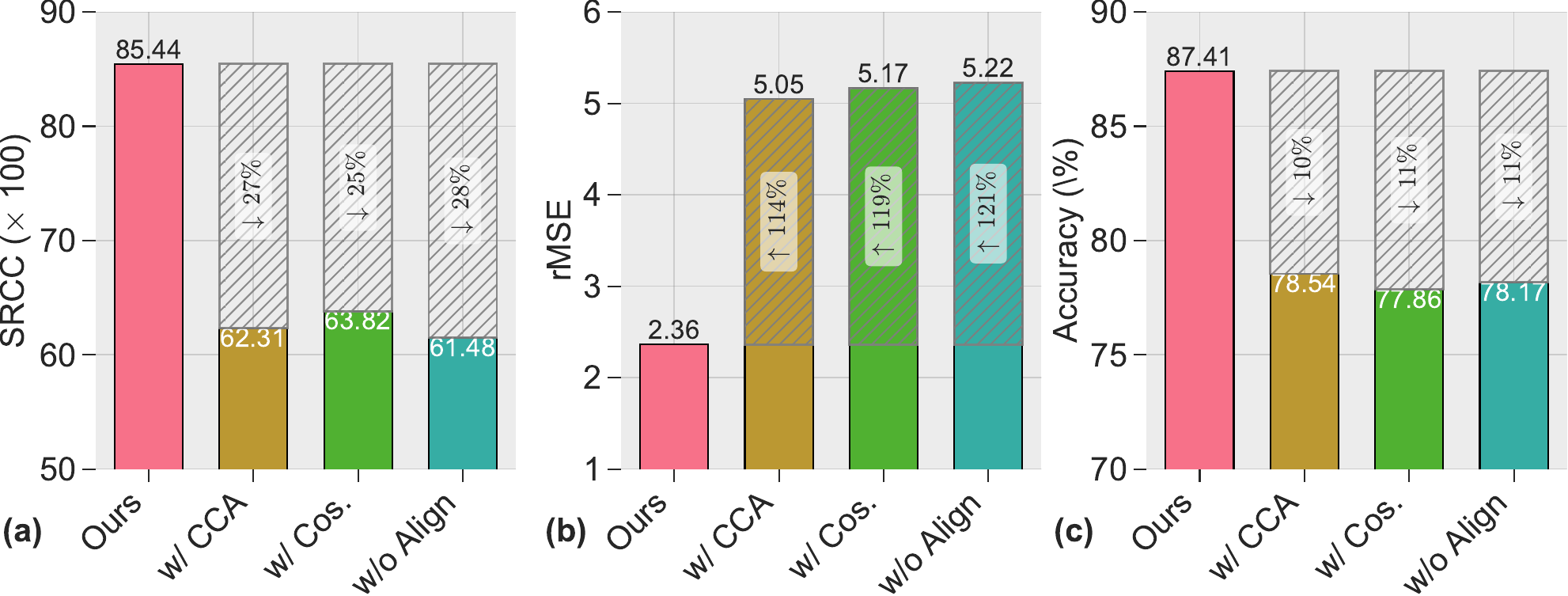}
    \caption{Ablation results of different alignment strategies on MM--JDM. (a) SRCC (b) rMSE, and (c) Accuracy. The shaded regions indicate relative performance degradation compared with the full model (Ours), where $\downarrow$ and $\uparrow$ denote relative decrease and increase, respectively.}{
    \label{fig:loss}
        \phantomsubcaption\label{fig:loss-a}
        \phantomsubcaption\label{fig:loss-b}
        \phantomsubcaption\label{fig:loss-c}
    }
\end{figure}

To understand the contributions of individual components in DualAlign, we conduct a comprehensive ablation study. The core results are presented in \cref{tab:ablation,fig:ablation-increment}. 
\cref{tab:ablation} follows a drop-one protocol relative to the full model, where each modality or design component is removed independently to assess its contribution within the complete system, rather than constructing the model incrementally from a minimal baseline.  In addition to these primary ablations, we further evaluate the effects of staged alignment design (see \cref{fig:stage}), different backbone architectures (see \cref{tab:backbone}), and the influence of each loss component on optimization dynamics (see \cref{fig:loss}). 

\myPara{Impact of Individual Modalities}
To assess the contribution of each sensing modality, we remove individual inputs from DualAlign and compare the resulting performance against the full model (see \cref{tab:ablation}). The results reveal that each modality contributes complementary information to the learned representation.
Removing the RGB video stream leads to the most severe degradation, with SRCC dropping by 45\%, rMSE increasing by 187\%, and accuracy decreasing by 23\%, indicating that video provides the primary source of spatiotemporal information. 
Skeleton and optical flow also play important roles. Removing skeleton features reduces SRCC by 29\% and increases rMSE by 122\%, while removing optical flow yields a 25\% SRCC reduction and a 118\% rMSE increase. Although motion cues can in principle be inferred from RGB videos, explicitly modeling them via optical flow provides more stable and disentangled temporal representations, especially under limited-data conditions. Similarly, skeleton features capture structured motion patterns that are difficult to recover directly from raw pixels.
Textual descriptions further enrich the representation by providing high-level semantic context. Eliminating text leads to a 13\% SRCC drop and a 175\% increase in rMSE, indicating its role in improving prediction stability when visual signals are ambiguous or incomplete.
Overall, these results indicate that each modality contributes distinct and complementary cues, and removing any modality disrupts the learned multi-modal representation.

\myPara{Why Alignment Matters Beyond Modality Accumulation}
To further examine whether performance gains come from simply adding modalities or from effective cross-modal alignment, we use CoFInAl \cite{zhou2024cofinal} as the unimodal baseline and progressively extend it from RGB input (V) to V+F, V+F+S, and V+F+S+T, before comparing it with the full DualAlign model. CoFInAl is selected because it is a representative strong unimodal AQA baseline built on RGB inputs.
As shown in \cref{fig:ablation-increment}, naive modality accumulation leads to only limited and unstable improvements. Although adding flow and skeleton slightly improves correlation, the regression error remains high and even increases when all modalities are fused without alignment, indicating that heterogeneous inputs introduce representation discrepancies rather than consistent benefits.
In contrast, DualAlign achieves substantial improvements across all metrics after applying the proposed discrepancy-aware alignment mechanism, with SRCC increasing from 63.82 to 85.44 and rMSE decreasing from 5.166 to 2.363. This demonstrates that the main performance gains come from principled cross-modal alignment rather than modality accumulation alone.

\begin{figure}
    \centering
    \includegraphics[width=\linewidth]{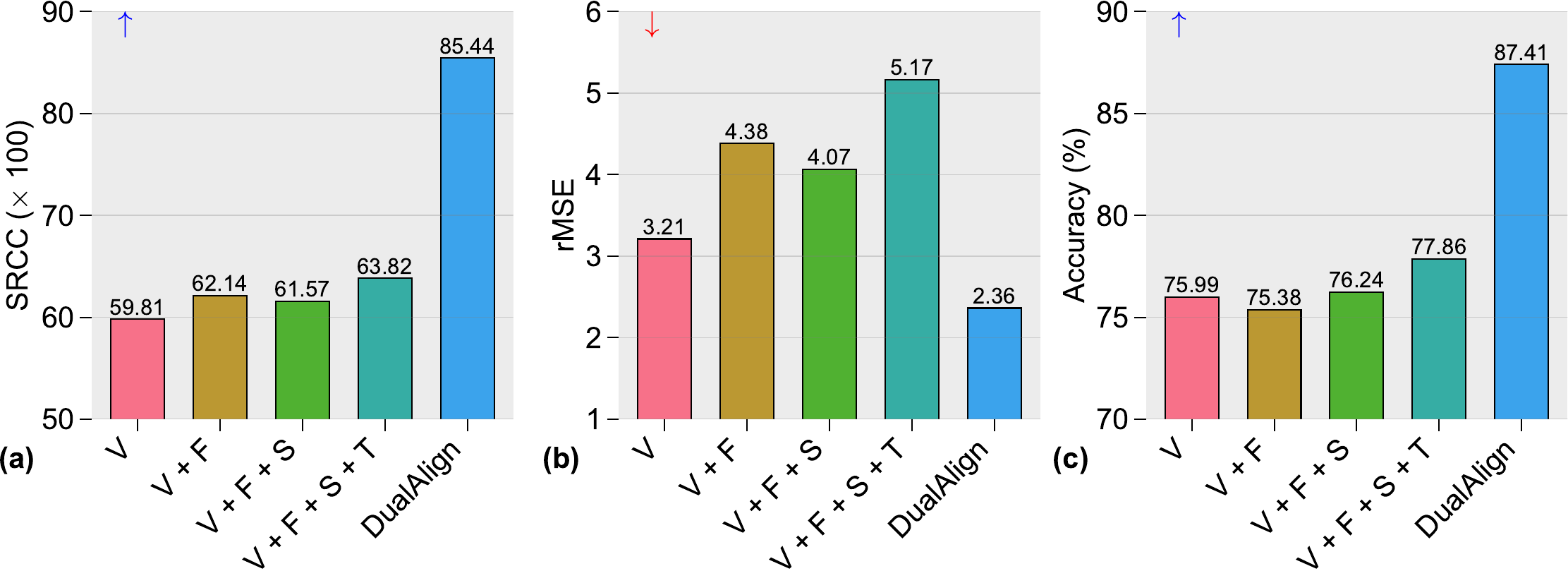}
    \MSfigcaption{fig:ablation-increment}{Progressive modality analysis from unimodal RGB input (V) to full multi-modal fusion. Simply adding modalities (Flow, Skeleton, and Text) provides limited and unstable improvements, while the proposed discrepancy-aware alignment in DualAlign leads to substantial gains across all metrics.}{}
\end{figure} 

\myPara{Impact of Alignment Mechanisms}
We further analyze the role of each alignment component by selectively removing them and evaluating the resulting performance (see \cref{tab:ablation}). The results show that the main performance gain does not come from directly applying the original GRAM loss, but from redesigning alignment to match the progressive fusion process.
Removing the visual feature alignment loss $\mathcal{L}_{\mathrm{vfs}}$ leads to a 27\% drop in SRCC and a 106\% increase in rMSE, indicating that aligning visual modalities is important for constructing a coherent representation. Replacing the proposed GRAM-based alignment with cosine similarity further degrades performance, confirming that geometric alignment captures inter-modal relationships more effectively than simple pairwise similarity.
More importantly, replacing the proposed two-stage alignment with the original single-stage Gram-based alignment results in a 29\% reduction in SRCC and a 104\% increase in rMSE. Notably, this one-stage Gram-based variant performs even worse than removing alignment entirely, indicating that directly applying a naive single-stage GRAM loss is insufficient and may even harm performance when modalities interact at different semantic levels.
This demonstrates that the improvement does not come from the GRAM loss itself, but from redesigning it into a discrepancy-aware two-stage alignment mechanism that is compatible with hierarchical visual-to-text fusion. Removing either stage individually also leads to noticeable degradation, further showing that the staged design provides complementary benefits for stabilizing cross-modal fusion.
Overall, these results show that alignment and architecture must work together: naive single-stage alignment is insufficient, while the proposed discrepancy-aware two-stage design effectively reduces cross-modal representation gaps. The relatively large performance drops observed in \cref{tab:ablation} reflect the tightly coupled design of DualAlign, where each component addresses a specific aspect of cross-modal representation discrepancies. As a result, removing any single component leads to noticeable degradation in the overall representation quality. 

\begin{table}
    \centering
    \caption{Ablation results on our MM--JDM dataset. Reported percentages indicate the change in performance relative to the full model (first row).}
    \setlength{\tabcolsep}{6pt}
    \resizebox{\linewidth}{!}{
        \begin{tabular}{llll}
            \toprule
            \textbf{Setting}                                                                                      & \textbf{Avg. SRCC}                & \textbf{Avg. rMSE}                 & \textbf{Avg. Acc.}                \\
            \midrule
            \rowcolor{orange!8} Ours                                                                              & \textbf{85.44}                    & \textbf{2.363}                     & \textbf{87.41}                    \\
            \rowcolor{yellow!8} \quad w/o Alignment (w/o $\mathcal{L}_{\mathrm{vfs}},\mathcal{L}_{\mathrm{vit}}$) & 63.82$^{\,\downharpoonright25\%}$ & 5.166$^{\,\downharpoonright119\%}$ & 77.86$^{\,\downharpoonright11\%}$ \\
            \rowcolor{yellow!8} \quad w/o GRAM (w/ Cosine Similarity)                                             & 61.48$^{\,\downharpoonright28\%}$ & 5.225$^{\,\downharpoonright121\%}$ & 78.17$^{\,\downharpoonright11\%}$ \\
            \rowcolor{yellow!8} \quad w/o Two-stage Alignment (w/ One-stage)                                      & 60.55$^{\,\downharpoonright29\%}$ & 4.819$^{\,\downharpoonright104\%}$ & 77.97$^{\,\downharpoonright11\%}$ \\
            \rowcolor{yellow!8} \quad w/o $\mathcal{L}_{\mathrm{vfs}}$ (w/ Second-stage)                          & 62.77$^{\,\downharpoonright27\%}$ & 4.858$^{\,\downharpoonright106\%}$ & 78.47$^{\,\downharpoonright10\%}$ \\
            \rowcolor{yellow!8} \quad w/o $\mathcal{L}_{\mathrm{vit}}$ (w/ First-stage)                           & 62.38$^{\,\downharpoonright27\%}$ & 4.958$^{\,\downharpoonright110\%}$ & 77.54$^{\,\downharpoonright11\%}$ \\
            \rowcolor{brown!8} \quad w/o Text Modality                                                            & 74.50$^{\,\downharpoonright13\%}$ & 6.508$^{\,\downharpoonright175\%}$ & 74.76$^{\,\downharpoonright14\%}$ \\
            \rowcolor{brown!8} \quad w/o Video Modality                                                           & 46.91$^{\,\downharpoonright45\%}$ & 6.782$^{\,\downharpoonright187\%}$ & 66.95$^{\,\downharpoonright23\%}$ \\
            \rowcolor{brown!8} \quad w/o Flow Modality                                                            & 63.67$^{\,\downharpoonright25\%}$ & 5.155$^{\,\downharpoonright118\%}$ & 78.84$^{\,\downharpoonright10\%}$ \\
            \rowcolor{brown!8} \quad w/o Skeleton Modality                                                        & 60.66$^{\,\downharpoonright29\%}$ & 5.244$^{\,\downharpoonright122\%}$ & 75.78$^{\,\downharpoonright13\%}$ \\
            \bottomrule
        \end{tabular}
    }
    \label{tab:ablation}
\end{table}

\myPara{Impact of the Two-Stage Strategy}
We further examine the effectiveness of the proposed two-stage fusion strategy by comparing it with several representative one-stage variants, including concatenation fusion, additive fusion, and cross-attention fusion. Although the cross-attention variant performs multiple attention interactions across modalities, all operations occur within a single fusion block and therefore still constitute a one-stage mechanism under standard multi-modal learning practice.
As shown in \cref{fig:stage}, the performance differences across methods are substantial. Relative to the two-stage DualAlign (SRCC 85.44), one-stage fusion variants exhibit pronounced degradation: concatenation drops by 59\%, additive fusion drops by 65\%, and cross-attention drops by 53\%. Even the one-stage Gram-based variant achieves only 60.55 SRCC, representing a 29\% reduction. Accuracy follows the same trend, with decreases of 38\%, 44\%, 30\%, and 11\% for the same methods, respectively. rMSE also worsens significantly, increasing by 150\% (concat), 201\% (additive), 111\% (cross-attention), and 104\% (one-stage Gram).
These results highlight a consistent pattern: single-step fusion lacks sufficient alignment capacity because early integration forces highly heterogeneous modalities into a shared embedding space, resulting in unstable optimization and weakened modality-specific cues. In contrast, the two-stage DualAlign framework first constructs a coherent visual representation from RGB, flow, and skeleton modalities, and then aligns this stabilized manifold with textual information. This staged decomposition leads to more stable training dynamics and yields consistently superior performance across all metrics.

\myPara{Impact of Fusion Order}
We further investigate whether the order in which modalities are fused affects overall performance. As shown in \cref{fig:stage}, reversing the fusion order, where textual features are injected into the RGB stream before alignment with optical flow and skeleton modalities, results in substantial performance degradation. Compared with the full two-stage DualAlign, which achieves an SRCC of 85.44 and an Accuracy of 87.41, the reverse variant attains only 73.12 SRCC and 82.35 Accuracy. This corresponds to a 14\% reduction in SRCC and a 6\% decrease in Accuracy. Meanwhile, rMSE increases from 2.363 to 3.487, representing a 48\% deterioration.
These results indicate that the premature introduction of textual semantics disrupts the formation of a coherent visual manifold, thereby weakening subsequent alignment with other visual modalities. In contrast, fusing visual modalities first and introducing textual information only after visual stabilization yields consistently superior performance. This confirms that the fusion order is not interchangeable and that preserving visual coherence prior to cross-modal integration is crucial for stable optimization and accurate prediction.

\begin{table}[!t]
    \centering
    \scriptsize
    \MStabcaption{tab:backbone}{Comparison results with different backbones on the MM--JDM dataset.}
    \renewcommand{\arraystretch}{1.3} \setlength{\tabcolsep}{2pt}
    \resizebox{\linewidth}{!}{
        \begin{tabular}{c
            >{\centering\arraybackslash}m{0.33\linewidth}
            >{\centering\arraybackslash}m{0.33\linewidth}
            >{\centering\arraybackslash}m{0.33\linewidth}}
            \toprule
             & \bf CLIP \cite{radford2021learning} (Ours)
             & \bf MedCLIP-ResNet \cite{wang2022medclip}
             & \bf MedCLIP-ViT \cite{wang2022medclip}                                                                                 \\
            \midrule

            \rowcolor{orange!8}
            \parbox[c][2.9cm][c]{0.4cm}{\centering\rotatebox{90}{Similarity Distribution}}
             & \includegraphics[width=2.9cm,clip]{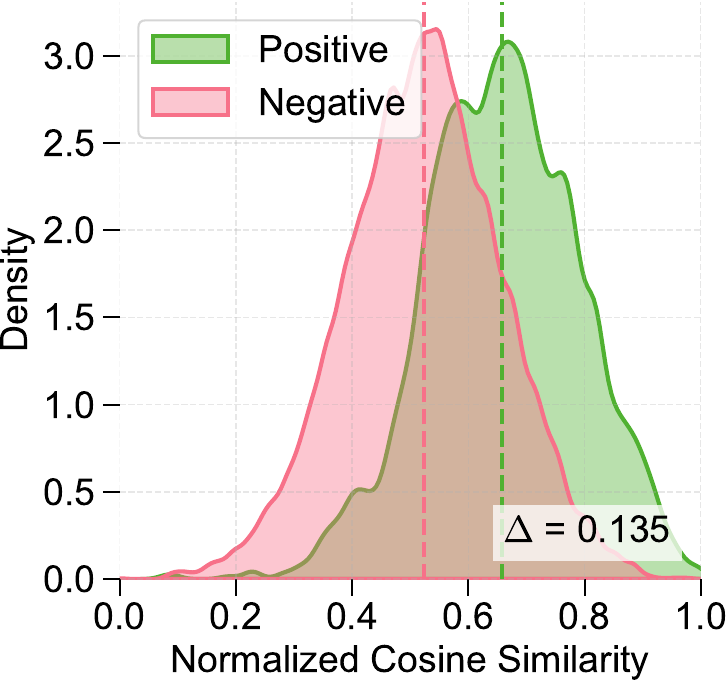}        & \includegraphics[width=2.9cm,clip]{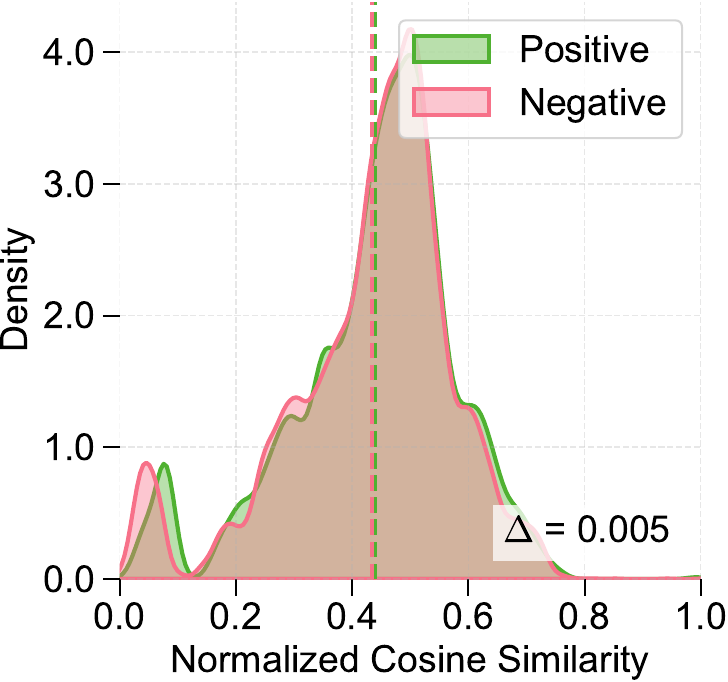}
             & \includegraphics[width=2.9cm,clip]{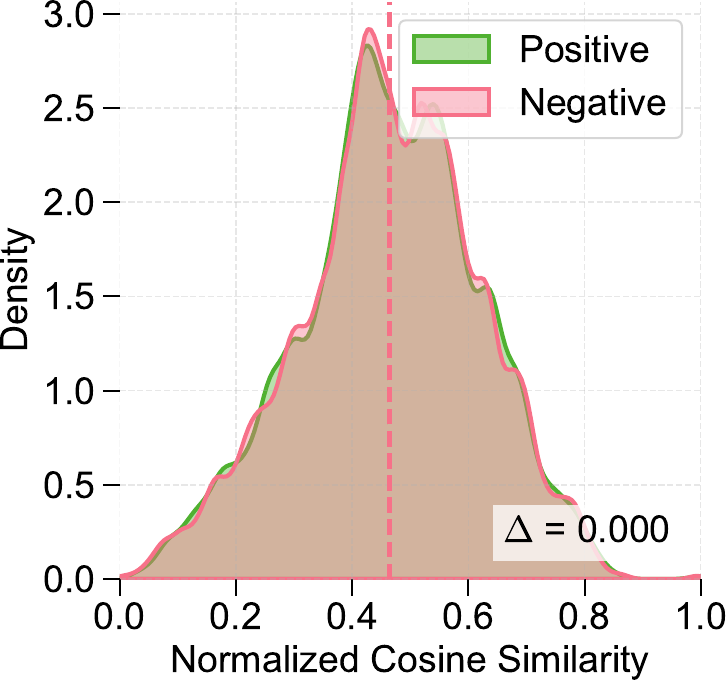}                                                               \\

            \rowcolor{yellow!8}
            \parbox[c][1.0cm][c]{0.4cm}{\centering\rotatebox{90}{Metric~~~}}
             & $\begin{array}{r@{\makebox[2em][c]{=}}l}
                        \text{SRCC}     & \text{0.8544}  \\
                        \text{rMSE}     & \text{2.363}   \\
                        \text{Accuracy} & \text{87.41\%} \\
                    \end{array}$
             & $\begin{array}{r@{\makebox[2em][c]{=}}l}
                        \text{SRCC}     & \text{0.8203}  \\
                        \text{rMSE}     & \text{2.874}   \\
                        \text{Accuracy} & \text{84.48\%} \\
                    \end{array}$
             & $\begin{array}{r@{\makebox[2em][c]{=}}l}
                        \text{SRCC}     & \text{0.8105}  \\
                        \text{rMSE}     & \text{3.445}   \\
                        \text{Accuracy} & \text{82.14\%} \\
                    \end{array}$                                                                          \\
            \bottomrule
        \end{tabular}
    }
\end{table}

\myPara{Impact of Different Backbones}
We further compare different video–text backbones to assess how pretraining domains affect multi-modal alignment on MM--JDM (see \cref{tab:backbone}). Our dataset contains everyday human-motion videos paired with concise action-focused descriptions, which differ substantially from the radiology images and disease-centric narratives used to pretrain medical-domain models \cite{wang2022medclip,yang2026unipet,zhou2026sdpt,yang2025restore_rwkv,yang2024amir,wu2024attriprompter}. This mismatch leads to pronounced differences in alignment quality.
The similarity-distribution plots in the first row of \cref{tab:backbone} show that CLIP \cite{radford2021learning} produces a clear separation between positive and negative video–text pairs ($\Delta = 0.135$), whereas MedCLIP-ResNet exhibits only marginal separation ($\Delta = 0.005$) and MedCLIP-ViT collapses to complete overlap ($\Delta = 0.000$). This indicates that CLIP is substantially better at distinguishing matched from mismatched pairs, while MedCLIP \cite{wang2022medclip} suffers from severe domain mismatch.
The quantitative results further validate this trend. Relative to MedCLIP-ResNet, CLIP improves SRCC from 0.8203 to 0.8544 (+4.2\%), reduces rMSE by 17.8\% (2.874 $\rightarrow$ 2.363), and increases Accuracy by 3.5 percentage points (84.48\% $\rightarrow$ 87.41\%). Compared with MedCLIP-ViT, the improvements are even larger: SRCC increases from 0.8105 to 0.8544 (+5.4\%), rMSE decreases by 31.4\% (3.445 $\rightarrow$ 2.363), and Accuracy improves by 6.4 percentage points (82.14\% $\rightarrow$ 87.41\%).
These consistent gains confirm that general-domain vision–language pretraining (CLIP) transfers effectively to action-oriented multi-modal assessment, whereas medical-domain pretraining is ill-suited for modeling skeletal dynamics, optical flow, and concise behavioral descriptions, leading to degraded alignment and prediction performance on MM--JDM.

\myPara{Impact of Alignment Losses}
As shown in \cref{fig:loss-a,fig:loss-b,fig:loss-c}, we compare different alignment strategies under a controlled setting in terms of SRCC, rMSE, and Accuracy, respectively. Removing the alignment module (\textit{w/o Alignment}) causes a clear and consistent degradation across all three metrics: the average SRCC and Accuracy drop by more than 25\% and 10\%, respectively (see \cref{fig:loss-a,fig:loss-c}), while the average rMSE more than doubles (see \cref{fig:loss-b}). This confirms that explicit cross-modal alignment is indispensable for reliable multi-modal assessment. Replacing the proposed GRAM-based alignment with simpler correlation-based objectives, such as cosine similarity and CCA, further reduces performance. In particular, the cosine-based variant yields the largest rMSE increase of about 120\% (see \cref{fig:loss-b}) together with a substantial SRCC decrease (see \cref{fig:loss-a}), indicating that pairwise similarity alone cannot preserve the higher-order relational structure among modalities required for fine-grained regression. CCA behaves similarly to cosine similarity and remains consistently below the GRAM-based model on all three metrics (see \cref{fig:loss}). Overall, the full DualAlign configuration achieves the best trade-off among correlation, regression precision, and discrete grading accuracy, which justifies the use of GRAM as the primary alignment objective.

\subsection{Quantitative and Qualitative Analysis} \label{sec:qq}
\begin{figure}[!t]
    \centering
    \sf
    \begin{overpic}[width=\linewidth]{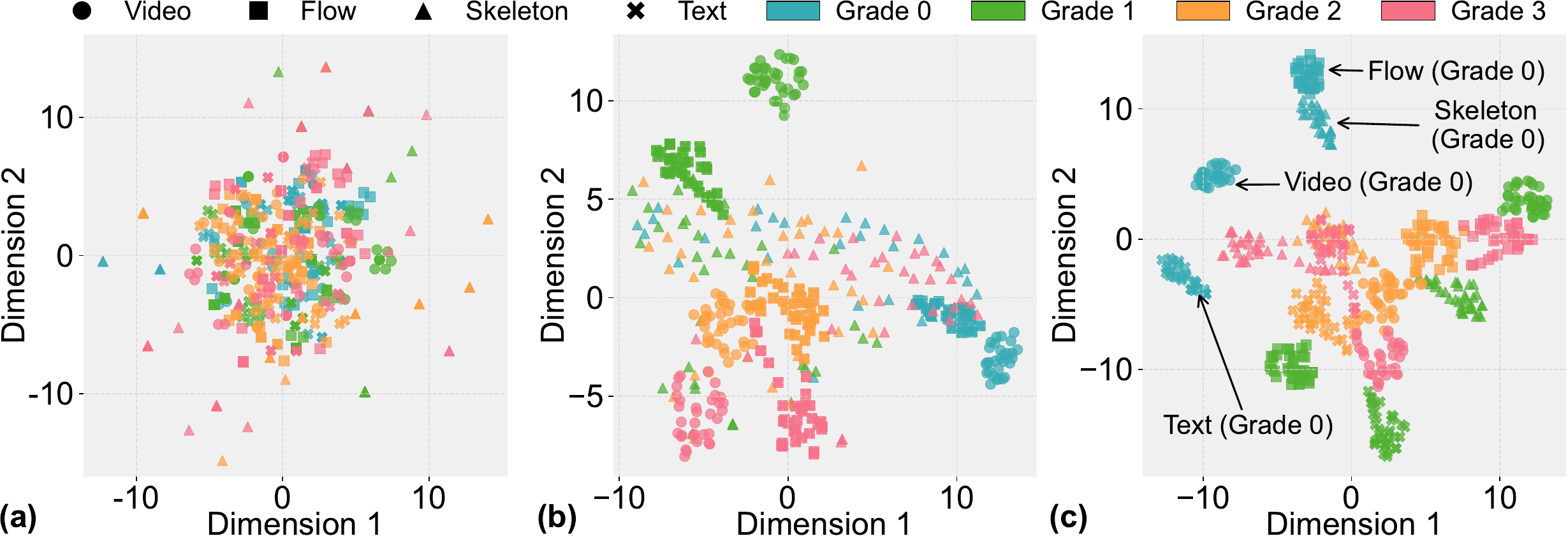}
    \end{overpic}
    \caption{t-SNE visualization of different modality embeddings. (a) shows the embeddings of Video, Flow, Skeleton, and Text modalities, (b) shows the embeddings without DualAlign's alignment strategy, and (c) shows the embeddings after applying DualAlign's alignment strategy.}
    \label{fig:tsne-plot}{
        \phantomsubcaption\label{fig:tsne-a}
        \phantomsubcaption\label{fig:tsne-b}
        \phantomsubcaption\label{fig:tsne-c}
    }
\end{figure}

\begin{figure}
    \centering
    \includegraphics[width=\linewidth]{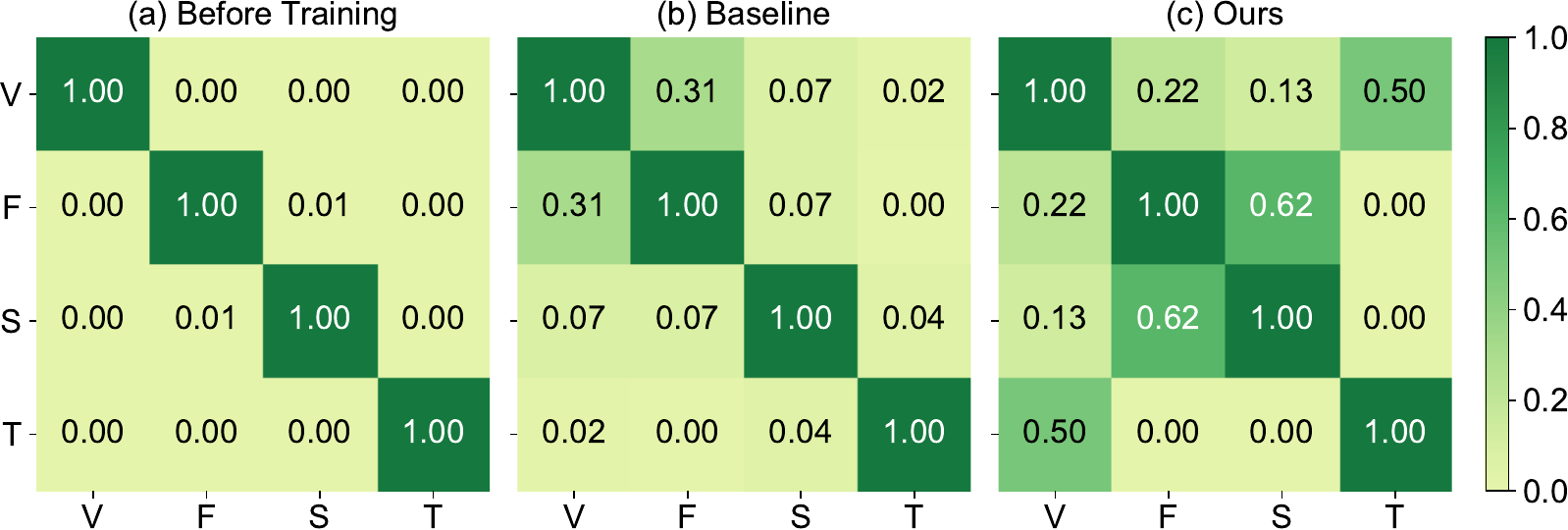}
    \caption{CKA similarity heatmaps between Video (V), Flow (F), Skeleton (S), and Text (T) shown in (a) the initial pretrained state, (b) after one-stage concatenation fusion, and (c) after our two-stage alignment.
        (a) shows that the four modalities occupy largely uncorrelated feature spaces.
        (b) indicates that one-stage fusion produces only partial and unstable alignment.
        (c) demonstrates that our method yields coherent cross-modal alignment.}{
        \label{fig:cka}
        \phantomsubcaption\label{fig:cka-a}
        \phantomsubcaption\label{fig:cka-b}
        \phantomsubcaption\label{fig:cka-c}
    }
\end{figure}

\begin{figure}
    \centering
    \includegraphics[width=\linewidth]{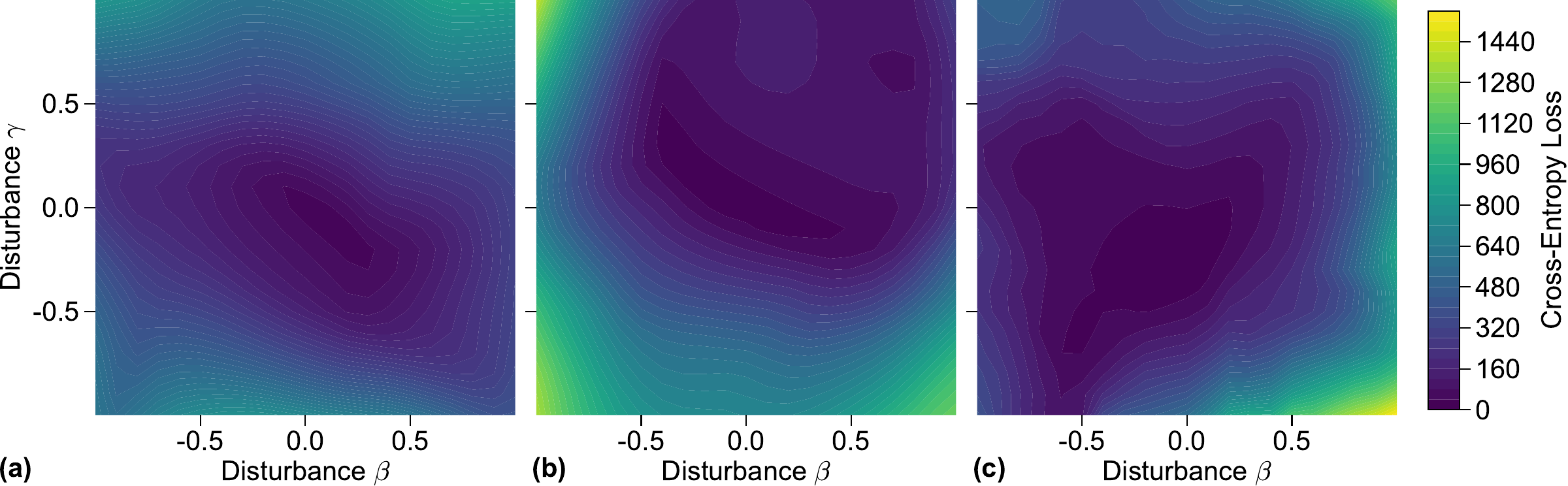}
    \caption{Two-dimensional loss landscapes obtained by random parameter perturbations around the converged solutions for different fusion strategies, illustrating the local smoothness and optimization stability.
        (a) DualAlign without alignment,
        (b) DualAlign with cosine-similarity alignment, and
        (c) the full DualAlign model with GRAM-based alignment.}{
        \label{fig:landscape}
        \phantomsubcaption\label{fig:landscape-a}
        \phantomsubcaption\label{fig:landscape-b}
        \phantomsubcaption\label{fig:landscape-c}
    }
\end{figure}

\begin{figure}[!t]
    \centering
    \sf
    \begin{overpic}[width=\linewidth]{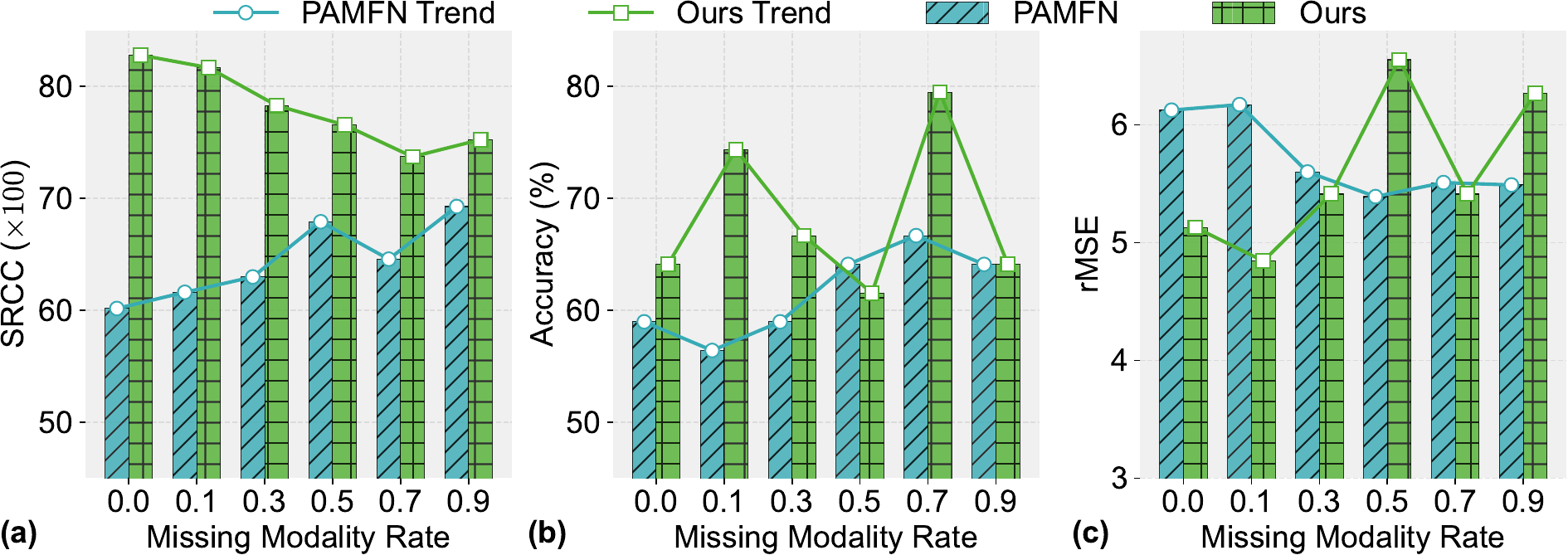}
    \end{overpic}
    \caption{Performance comparison with the state-of-the-art PAMFN method \cite{zeng2024multimodal} under the missing modality settings on the MM--JDM dataset. The $x$-axis represents the missing modality rate, and the $y$-axis indicates (a) SRCC, (b) Accuracy, and (c) rMSE, respectively.}
    \label{fig:miss-modal-plot}{
        \phantomsubcaption\label{fig:miss-modal-plot-a}
        \phantomsubcaption\label{fig:miss-modal-plot-b}
        \phantomsubcaption\label{fig:miss-modal-plot-c}
    }
\end{figure}

\begin{figure}[!t]
    \centering
    \sf
    \begin{overpic}[width=\linewidth]{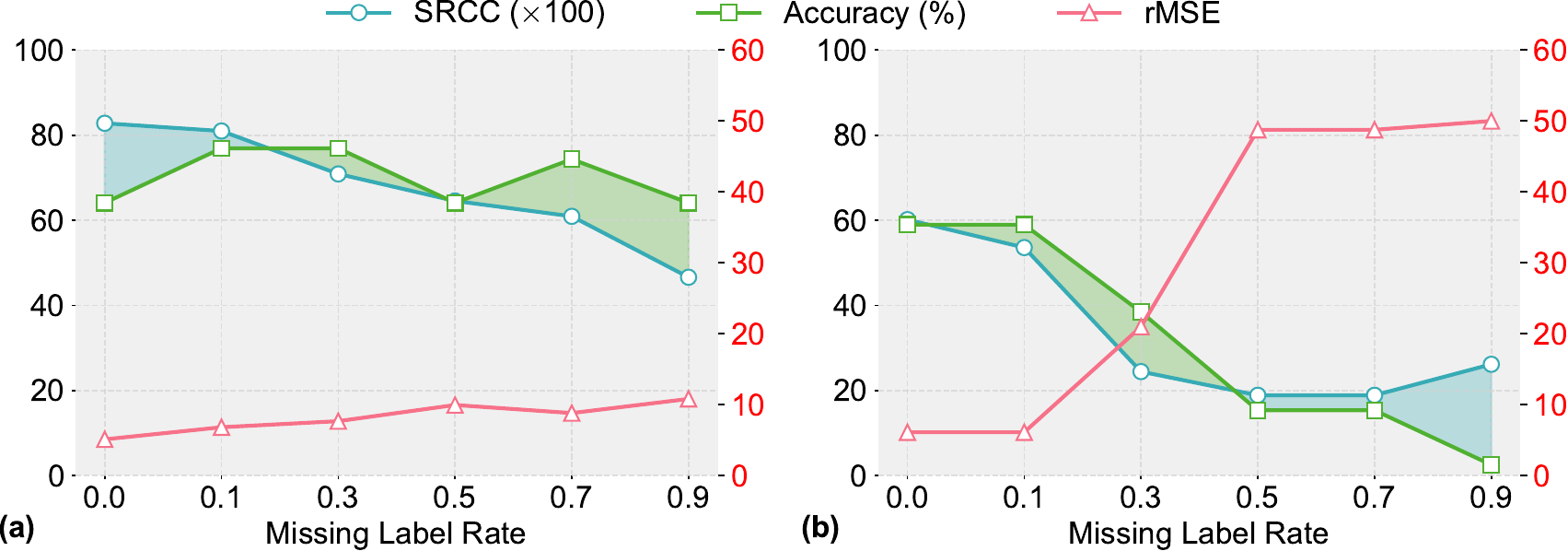}
    \end{overpic}
    \caption{Performance under the missing label settings on the MM--JDM dataset: (a) Our DualAlign method, (b) PAMFN \cite{zeng2024multimodal}. The $x$-axis represents the missing label rate, and the $y$-axis indicates the performance including SRCC (left), Accuracy (left), and rMSE (right).}
    \label{fig:miss-label-plot}{
        \phantomsubcaption\label{fig:miss-label-plot-a}
        \phantomsubcaption\label{fig:miss-label-plot-b}
    }
\end{figure}

\begin{figure}[!t]
    \centering
    \includegraphics[width=\linewidth]{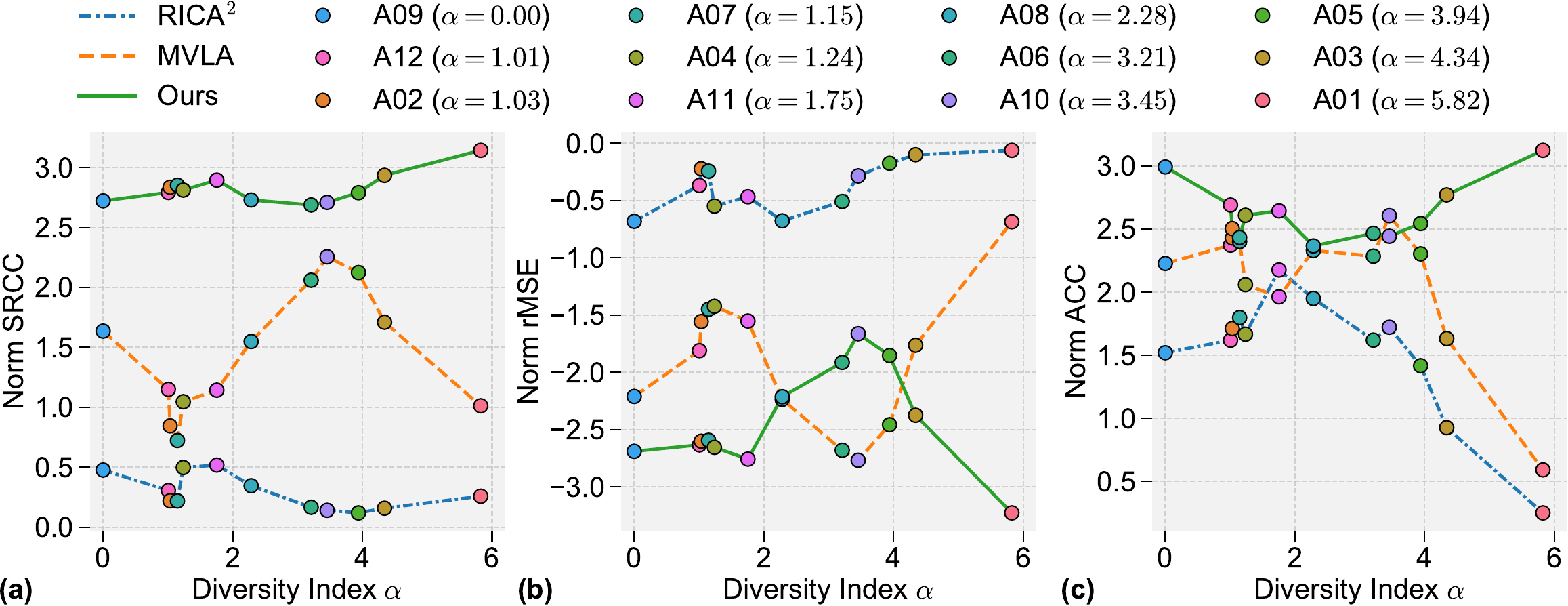}
    \caption{
        Robustness analysis under action-grade imbalance using the diversity index $\alpha$.
        (a) Normalized SRCC, (b) normalized rMSE, and (c) normalized Accuracy plotted against the diversity levels of all 12 actions from the MM--JDM dataset.
    }
    \label{fig:diversity}
    \phantomsubcaption\label{fig:diversity-a}
    \phantomsubcaption\label{fig:diversity-b}
    \phantomsubcaption\label{fig:diversity-c}
\end{figure}

\begin{figure*}
    \centering
    \includegraphics[clip,trim=25 230 25 220,width=\linewidth]{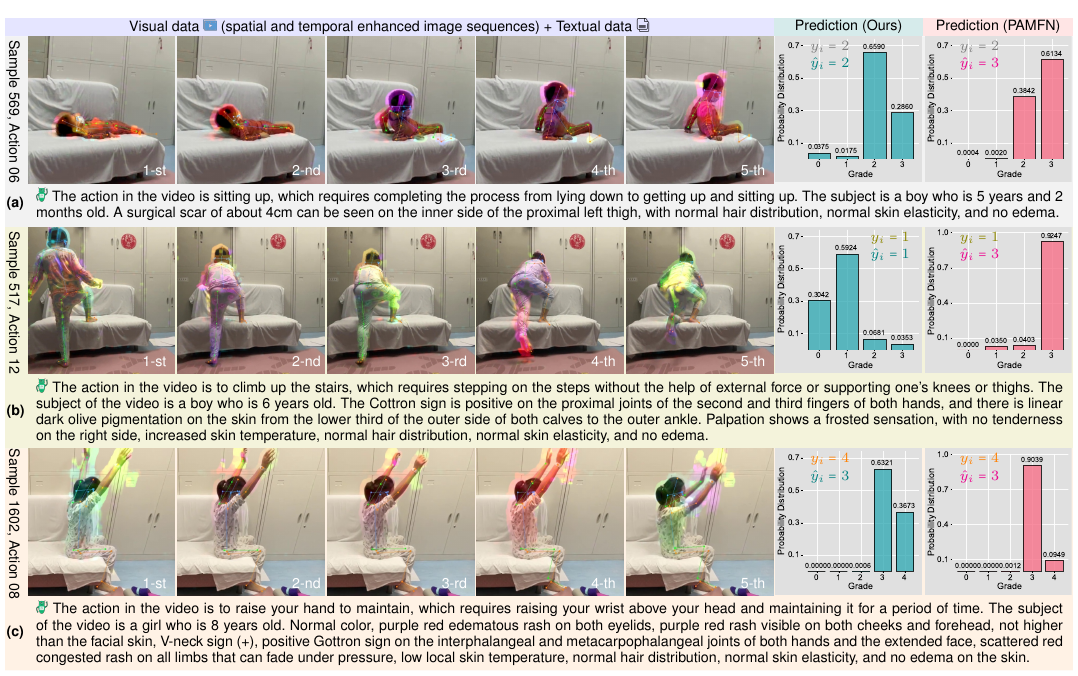}
    \caption{
        Three samples and their predicted grade distributions on MM--JDM: the first five columns show different frames with the visualized skeleton and flow. The last two columns display the predicted grade distribution plots of our method and the PAMFN method \cite{zeng2024multimodal}.
    }
    \label{fig:case_study}{
        \phantomsubcaption\label{fig:case_study-a}
        \phantomsubcaption\label{fig:case_study-b}
        \phantomsubcaption\label{fig:case_study-c}
    }
\end{figure*}

This section provides a comprehensive evaluation of the alignment effects and robustness of DualAlign. We first visualize feature distributions using t-SNE to illustrate inter-modal separability (see \cref{fig:tsne-plot}) and employ CKA similarity analysis to quantify modality alignment (see \cref{fig:cka}). We then examine the optimization behavior by analyzing the loss landscape (see \cref{fig:landscape}). Beyond feature-level insights, we evaluate DualAlign under several real-world conditions, including missing-modality scenarios (see \cref{fig:miss-modal-plot}), label-scarce regimes (see \cref{fig:miss-label-plot}), and action-grade imbalance (see \cref{fig:diversity}). Finally, we include a case study (see \cref{fig:case_study,fig:case_study-rg}) to qualitatively validate the effectiveness of the proposed two-stage fusion strategy in multi-modal action assessment.

\myPara{Visualization of Alignment Effect}
To visualize the alignment effectiveness of our DualAlign method, we conduct a t-SNE analysis on the embeddings of different modalities before and after applying DualAlign. The results are presented in \cref{fig:tsne-plot}. The t-SNE visualization shows the embeddings of Video, Flow, Skeleton, and Text modalities. \cref{fig:tsne-a} illustrates the embeddings of different modalities, highlighting the distinct clusters and separations between modalities. \cref{fig:tsne-b} shows the embeddings without DualAlign's alignment strategy, where the modalities are not well-aligned, leading to overlapping clusters and reduced separations. In contrast, \cref{fig:tsne-c} shows the embeddings after applying DualAlign's alignment strategy, where the modalities are effectively aligned, resulting in clear clusters and enhanced separations.
The alignment effect is particularly evident in the embeddings of the same grade samples and modalities, which are more clustered and separated from other grades compared to \cref{fig:tsne-b}, indicating improved alignment and fusion across modalities.
As corresponding to the results in \cref{tab:ablation}, the alignment mechanisms in our DualAlign method play a critical role in enhancing model performance and robustness.
These results validate the effectiveness of our DualAlign method in capturing inter-modal dependencies and improving alignment capabilities, providing valuable insights into JDM-MSA.

\myPara{Cross-Modal Alignment Analysis}
To further understand why the proposed alignment strategy outperforms one-stage fusion, we visualize cross-modal representational similarity using Centered Kernel Alignment (CKA) in \cref{fig:cka}. All CKA values are computed by averaging cross-modal similarities across the twelve action classes of the MM--JDM dataset, providing a dataset-level view of alignment behavior.
As shown in \cref{fig:cka-a}, the four modalities (V, F, S, T) initially occupy nearly uncorrelated feature spaces, with off-diagonal similarities close to zero (e.g., V–F: $0.03$, V–S: $0.02$), confirming the severe cross-modal misalignment observed in practice. After applying a standard one-stage fusion strategy (\cref{fig:cka-b}), cross-modal similarities increase only modestly (e.g., V–F: $0.31$, V–S: $0.07$, F–S: $0.07$), and several modality pairs remain weakly correlated, indicating that one-stage fusion achieves only partial and unstable alignment.
In contrast, DualAlign produces a structured and balanced similarity pattern (\cref{fig:cka-c}). Visual modalities exhibit consistently high mutual similarity (e.g., V–F: $0.62$, V–S: $0.50$, F–S: $0.60$), reflecting effective alignment of shared spatiotemporal structure. At the same time, similarities involving the text modality remain moderate rather than saturated, indicating semantic coupling without forcing full representational collapse.
Importantly, excessively high cross-modal similarity is neither expected nor desirable in multi-modal action assessment, as different modalities encode complementary but non-identical information. Over-alignment would suppress modality-specific characteristics and reduce their discriminative value. The observed CKA pattern therefore reflects a desirable balance, where shared action-relevant information is aligned while modality diversity is preserved. This explains why naive one-stage fusion struggles to reconcile heterogeneous feature spaces, particularly between visual streams and textual semantics. By stabilizing visual representations before introducing text, DualAlign achieves effective alignment without sacrificing robustness under noisy or inconsistent modalities.

\myPara{Stability and Optimization Behavior}
To analyze the optimization geometry and robustness of different fusion strategies, we visualize the two-dimensional loss landscape around the converged model parameters. Let $\theta$ denote the trained parameters, and construct two random perturbation directions $d_1$ and $d_2$ by sampling Gaussian noise with the same shape as each parameter tensor and performing layer-wise $\ell_2$ normalization such that $\lVert d_k^{(l)} \rVert_2 \approx \lVert \theta^{(l)} \rVert_2$ for each layer $l$ and $k \in \{1,2\}$. We then evaluate the validation cross-entropy loss at perturbed parameters $\theta(\beta,\gamma) = \theta + \beta d_1 + \gamma d_2$ over a $21 \times 21$ grid with $\beta,\gamma \in [-1.0, 1.0]$. To obtain a stable estimate of the landscape, this procedure is repeated with 10 independently sampled direction pairs and the loss values are averaged at each grid point; the final surfaces are visualized as contour plots, where flatter and wider low-loss regions indicate better robustness and generalization.  The resulting two-dimensional loss landscapes are presented in \cref{fig:landscape}, The model without alignment (see \cref{fig:landscape-a}) exhibits sharp, narrow valleys and rapidly changing contour lines, which indicate a highly anisotropic and poorly conditioned loss surface. The cosine-based alignment (see \cref{fig:landscape-b}) mildly improves smoothness but still shows irregular ridges and local fluctuations. In contrast, the full DualAlign model (see \cref{fig:landscape-c}) presents a much flatter and more coherent basin with sparse and smoothly varying contour lines, suggesting that the GRAM-based alignment effectively regularizes the parameter space, reduces sensitivity to initialization and batch-level noise, and yields a well-behaved optimization landscape. In practice, we observe that the alignment loss decays rapidly in the early training stage and then stabilizes, while the grading loss continues to decrease steadily with monotonic convergence and no oscillation. These empirical results directly address concerns regarding numerical robustness and loss conflicts, and indicate that the alignment and grading objectives cooperate rather than compete during training.

\myPara{Robustness of Missing Modalities}
Missing modalities are common in real-world scenarios due to sensor failures, occlusions, or incomplete data collection \cite{zhou2026brima}. To evaluate the robustness of DualAlign under such conditions, we simulate missing-modality scenarios on the MM--JDM dataset with varying missing rates (0.0, 0.1, 0.3, 0.5, 0.7, and 0.9).
In this setting, the model is trained with all modalities, while at inference time, certain modalities (e.g., textual or auxiliary visual inputs) are randomly removed at the sample level and replaced with empty placeholders. The missing rate indicates the proportion of samples in which a given modality is unavailable, while the RGB video input is always preserved. This setting differs from the ablation study in \cref{tab:ablation}, where modalities are removed during both training and inference to assess their individual contributions. In contrast, here we focus on inference-time missing modalities, evaluating the model's robustness when trained with full modalities but tested with incomplete inputs. The results are presented in \cref{fig:miss-modal-plot}. DualAlign maintains stable performance across different missing rates, achieving consistently high SRCC (see \cref{fig:miss-modal-plot-a}), Accuracy (see \cref{fig:miss-modal-plot-b}), and low rMSE (see \cref{fig:miss-modal-plot-c}). Even at a 70\% missing rate, the model retains strong performance (SRCC $\times 100$ of 73.72, Accuracy of 79.49\%, and rMSE of 5.413), outperforming PAMFN~\cite{zeng2024multimodal} under the same setting.
These results demonstrate that, by learning cross-modal relationships during training, DualAlign remains robust when modalities are partially unavailable at inference time.

\myPara{Robustness of Label Scarcity}
Label scarcity is a common challenge in medical datasets, where certain actions or grades may have limited samples, leading to class imbalance and potential performance degradation. To evaluate the robustness of our DualAlign method under label scarcity settings, we conduct experiments with varying missing label rates (0.0, 0.1, 0.3, 0.5, 0.7, and 0.9) on the MM--JDM dataset. In our setting, the labels of certain actions are missing, simulating real-world scenarios where certain grades may be unavailable.
The results are presented in \cref{fig:miss-label-plot}.
Although the missing label degrades the model performance, our DualAlign method (see \cref{fig:miss-label-plot-a}) demonstrates robust performance across different missing label rates, achieving high SRCC, Accuracy, and rMSE values. Even with a missing label rate of 90\%, our method maintains an SRCC ($\times 100$) of 46.63, an Accuracy of 64.10\%, and an rMSE of 10.826, outperforming the state-of-the-art PAMFN model \cite{zeng2024multimodal} under similar settings.
In contrast, the PAMFN model (see \cref{fig:miss-label-plot-b}) exhibits a significant performance decrease as the missing label rate increases, demonstrating the fragility of sparse labeled data scenarios on model performance.
These results highlight the robustness of DualAlign under label scarcity scenarios of real-world applications.

\myPara{Robustness to Action Grade Imbalance}
The MM--JDM dataset exhibits substantial heterogeneity in grade distributions across actions, which may introduce estimation bias and impair reliability, especially for actions with sparse or skewed label coverage.
To quantify this effect in a principled manner, we adopt the entropy-normalized diversity index $\alpha$:
\begin{equation}
    \MSeqlabel{eq:diversity_index}
    \alpha = - \frac{q}{\log(q)} \sum_{i=1}^{N} p_i \log p_i,
\end{equation}
where $q$ denotes the number of grade categories and $p_i$ represents the empirical proportion of samples in the $i$-th grade. This formulation provides a well-justified scalar characterization of grade dispersion by combining Shannon entropy with a scale normalization factor, ensuring that larger values of $\alpha$ reflect more uniform and therefore more challenging grade distributions. As shown in the legend of \cref{fig:diversity}, actions with the highest diversity indices, such as Action~1 ($\alpha = 5.82$) and Action~3 ($\alpha = 4.34$) indeed correspond to weaker absolute performance in \cref{tab:exp}, which confirms the adverse effect of severe grade imbalance. In contrast, for actions with very low diversity, such as Action~9 and Action~12 ($\alpha = 0.00$ and $\alpha = 1.01$), our method achieves the strongest results, which demonstrates its effectiveness in handling imbalanced scenarios.
To further assess robustness, we examine the normalized relative performance as a function of $\alpha$, obtained by subtracting the mean and dividing by the standard deviation over multi-modal AQA methods. We also include RICA$^{2}$~\cite{majeedi2024rica} and MVLA~\cite{xu2024vision} as two of the most competitive multi-modal baselines for comparison.
As diversity increases, DualAlign maintains highly stable normalized SRCC (remaining within the $2.5$–$3.5$ interval) and exhibits a consistent rise in normalized accuracy (from approximately $2.2$ to $3.5$), while rMSE decreases steadily. In contrast, both baselines undergo pronounced degradation: for instance, the normalized accuracy of RICA$^{2}$ drops by more than $67\%$ from low- to high-diversity actions, with MVLA showing a similar downward trend (see \cref{fig:diversity-c}). Since MM--JDM reflects real clinical imbalance patterns, these results demonstrate that DualAlign is markedly more resilient to grade-distribution heterogeneity, in line with the imbalance-tolerant behavior previously attributed to ETF-based formulations~\cite{zhou2024cofinal}.

\myPara{Case Study}
To provide a more intuitive understanding of DualAlign, we present qualitative results and case studies across all three datasets, including MM--JDM, RG, and FIS-V, as shown in \cref{fig:case_study,fig:case_study-rg}. Here, we take \cref{fig:case_study} as an example for detailed analysis.
We show three representative samples from MM--JDM, each containing visual data (spatial and temporal enhanced image sequences) together with textual descriptions. The samples correspond to different actions, including ``Head Lift'' (see \cref{fig:case_study-a}), ``Step On'' (see \cref{fig:case_study-b}), and ``Hand Raise and Maintain'' (see \cref{fig:case_study-c}), with varying grades reflecting different levels of JDM severity and mobility impairment. The visual data captures movement dynamics and muscle strength, while the textual modality provides complementary diagnostic context for MSA.
The last two columns show the predicted grade distributions of our DualAlign method and the strong baseline PAMFN~\cite{zeng2024multimodal}, highlighting the differences in prediction behavior. DualAlign demonstrates high consistency with the ground-truth grades (grade 2 for \cref{fig:case_study-a} and grade 1 for \cref{fig:case_study-b}), accurately reflecting the patients’ movement capability across different actions. In contrast, PAMFN exhibits lower prediction accuracy and less stable confidence distributions.
In \cref{fig:case_study-c}, both methods fail to predict the correct grade 4, reflecting the difficulty of assessing complex actions with severe functional impairment. Nevertheless, DualAlign produces predictions that are more concentrated around the ground-truth grade, indicating stronger confidence and better robustness.
Overall, similar trends can also be observed on RG and FIS-V in \cref{fig:case_study-rg}, where DualAlign consistently provides more reliable score distributions than PAMFN. These qualitative results further demonstrate the effectiveness of DualAlign in robust multi-modal action assessment.

\begin{figure*}
    \centering
    \includegraphics[clip,trim=25 235 25 230,width=\linewidth]{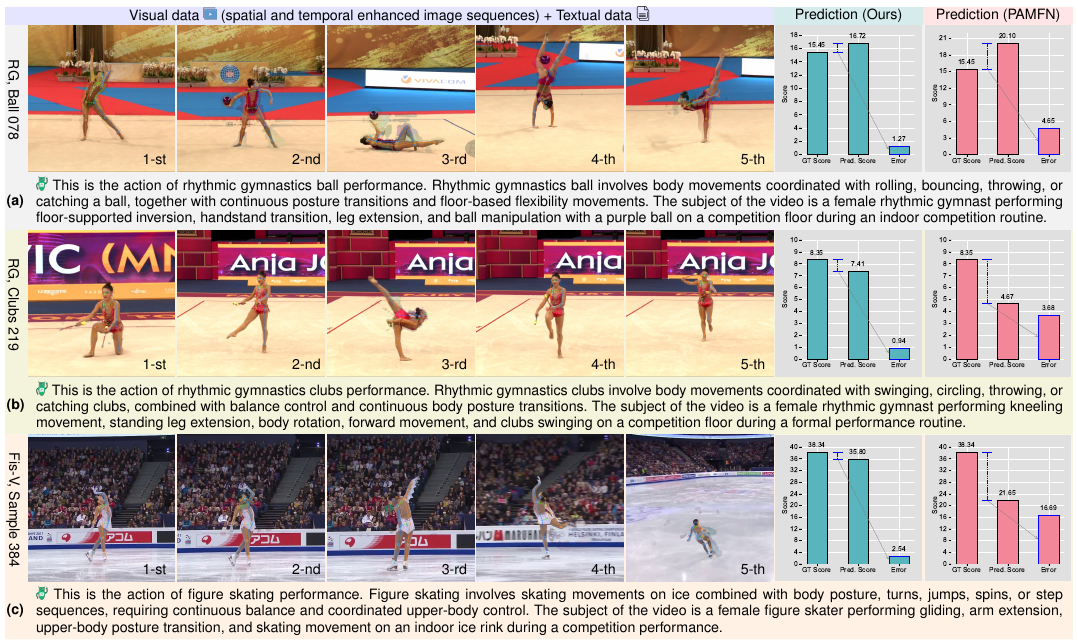}
    \MSfigcaption{fig:case_study-rg}{
        Representative samples and their predicted grade distributions on RG and Fis-V: the first five columns show different frames with the visualized skeleton and flow. The last two columns display the predicted score plots of our method and the PAMFN method \cite{zeng2024multimodal}.
    }{  
    \phantomsubcaption\label{fig:case_study-rg-a}
        \phantomsubcaption\label{fig:case_study-rg-b}
        \phantomsubcaption\label{fig:case_study-rg-c}
    }
\end{figure*}% 

\section{Conclusion and Discussion} \label{sec:conclusion}
In this work, we introduced DualAlign, a two-stage framework for multi-modal AQA that explicitly addresses the cross-modal representation discrepancies arising from heterogeneous inputs. The framework first constructs a stable and coherent visual representation by aligning RGB video, optical flow, and skeleton sequences, which together provide complementary appearance, motion, and structural cues. Textual semantics are incorporated only after the visual manifold has been consolidated, allowing high-level descriptions to enhance interpretation without interfering with early-stage visual feature formation. This staged design enables each modality to contribute effectively and reduces cross-modal interference.
We also presented MM--JDM, a comprehensive multi-modal dataset that integrates visual modalities with structured textual descriptions collected in real clinical environments. MM--JDM captures natural variability, noise, class imbalance, and label scarcity, offering a challenging and representative benchmark for evaluating multi-modal fusion and alignment strategies.
Extensive experiments demonstrate that DualAlign achieves state-of-the-art performance on MM--JDM and generalizes well to external AQA benchmarks. Ablation studies further validate the necessity of the two-stage alignment design and highlight the complementary benefits of integrating appearance, motion, structure, and semantics. These findings indicate that resolving representation discrepancies is essential for advancing robust multi-modal AQA systems.

\myPara{Limitation and Future Work}
Our work has several limitations. Several actions in the MM--JDM dataset exhibit substantial grade imbalance, as severe cases are naturally scarce in clinical practice. This limits the diversity of training samples and restricts generalization across different severity levels. Addressing this issue will require both improved data collection and more effective learning strategies, such as data augmentation and semi-supervised learning, to enhance robustness under limited-data conditions.
Although the proposed alignment strategy mitigates modality-specific noise, performance may still degrade under extreme occlusion or heavily corrupted visual inputs. This suggests the need for more robust modeling, motivating future research on occlusion-aware or uncertainty-guided approaches.
In addition, the use of fixed-length uniform sampling, while ensuring consistent temporal coverage and stable training, may lose fine-grained temporal details in actions with highly variable durations. Developing more flexible temporal modeling strategies, such as adaptive sampling or sliding-window approaches, could further improve the ability to capture subtle temporal variations, especially with larger-scale or more densely annotated data. Finally, while action-specific modeling enables fine-grained assessment under distinct scoring criteria, it may limit scalability in real-world scenarios where multiple actions need to be handled jointly. Extending the framework to unified multi-action models is therefore an important direction for improving flexibility and deployment efficiency. 

\section*{Acknowledgments}
We would like to express our sincere gratitude to Song Jin from Beijing Dianite Medical Technology Co., Ltd. for his support in data acquisition and processing. Correspondence may be addressed to Xiaohui Liang and Jianguo Li.

\section*{Data Availability Statement}
The MM--JDM dataset is available upon reasonable request, subject to ethical approval and data usage agreements. The source code for MM--JDM is publicly available at \url{https://github.com/Craaaaazy666/DualAlign}. The RG dataset is released by the official ACTION-NET repository at \url{https://github.com/qinghuannn/ACTION-NET}, and the FIS-V dataset is provided by its official benchmark implementation at \url{https://github.com/chmxu/MS_LSTM}.

\section*{Declarations}

\textbf{Ethical Approval:} 
This study was approved by the institutional ethics committee of the authors' institution.

\noindent\textbf{Consent to Participate:}
Informed consent was obtained from all participants or their legal guardians by the data providers.

\noindent\textbf{Consent for Publication:}
All authors have reviewed the manuscript and consent to its publication.

\noindent\textbf{Author Contributions:} Kanglei Zhou contributed to conceptualization, methodology, investigation, formal analysis, and writing the original draft. Ruizhi Cai contributed to methodology, software, validation, writing (review and editing), and data curation. Xinning Wang contributed to data curation, investigation, and resources. Yijian Zheng and Liyuan Wang contributed to writing (review and editing). Jianguo Li contributed to resources, supervision, and data acquisition. Xiaohui Liang contributed to supervision, project administration, and funding acquisition.

\noindent\textbf{Funding:}
This work was supported by the National Natural Science Foundation of China (No. 62272019) and the China Postdoctoral Science Foundation (No. 2025M781489).

\noindent\textbf{Conflict of Interest:} The authors declare that they have no conflicts of interest.

\bibliographystyle{spbasic}      
\bibliography{refs}

\end{document}